\documentclass[10pt,twocolumn,letterpaper]{article}

\usepackage[pagenumbers]{cvpr} 

\usepackage{algorithm}
\usepackage[accsupp]{axessibility}
\usepackage{algpseudocode}
\usepackage{bm}
\usepackage{booktabs}
\usepackage{mathtools}
\usepackage{url}

\definecolor{cvprblue}{rgb}{0.21,0.49,0.74}
\usepackage[pagebackref,breaklinks,colorlinks,citecolor=cvprblue]{hyperref}

\def\paperID{*****} 
\def\confName{CVPR}
\def\confYear{2025}

\title{StableAvatar: Infinite-Length Audio-Driven Avatar Video Generation}

\author{Shuyuan Tu$^{1}$ \ \ 
Yueming Pan$^{3}$ \ \ 
Yinming Huang$^{1}$ \ \ 
Xintong Han$^4$ \ \ 
Zhen Xing$^1$ \ \ 
Qi Dai$^{2}$ \ \ 
Chong Luo$^{2}$ \ \ \vspace{-0.01cm}\\
Zuxuan Wu$^1$ \ \ 
Yu-Gang Jiang$^{1}$ \\
{$^1$Fudan University}  \quad  
{$^2$Microsoft Research Asia}  \quad  
{$^3$Xi'an Jiaotong University}  \quad  
{$^4$Hunyuan, Tencent Inc.} \\
{\url{https://francis-rings.github.io/StableAvatar}}
}

\begin{document}

\twocolumn[{
\maketitle
\vspace{-3.2em}
\renewcommand\twocolumn[1][]{#1}
\begin{center}
    \centering
    \includegraphics[width=1\textwidth]{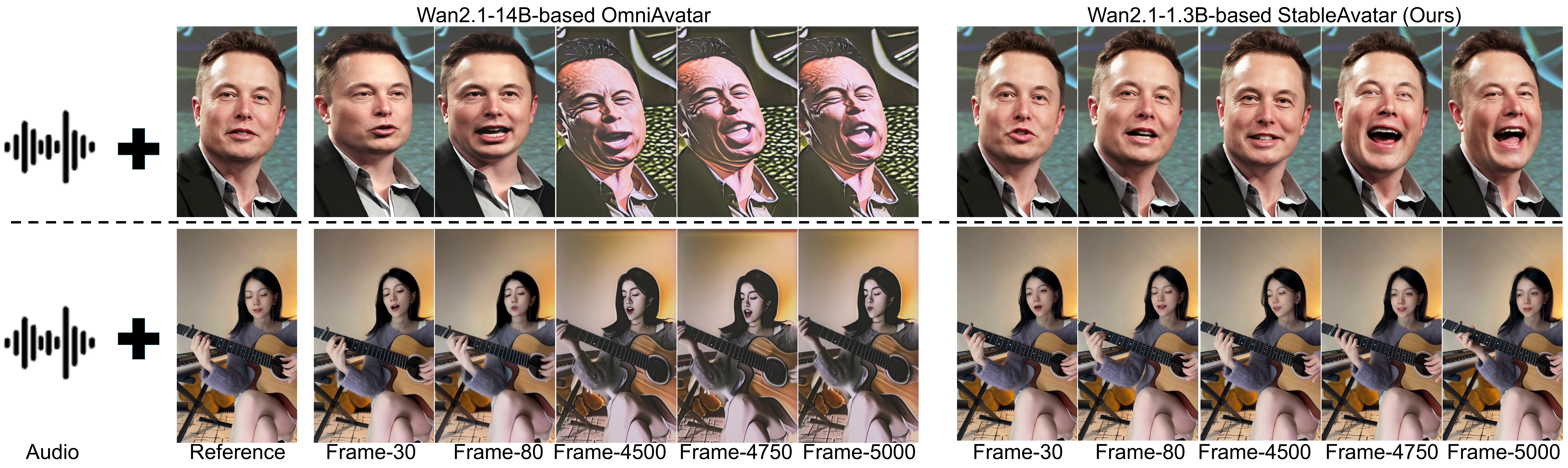}
    \vspace{-0.7cm}
    \captionof{figure}{Audio-driven avatar videos generated by StableAvatar, showing its power to synthesize infinite-length ID-preserving videos. The presented videos last over 3 minutes (FPS=30). Frame-X refers to the X-th frame of the synthesized avatar video.
    }
    \label{fig:cover}
\end{center}
}
]

\begin{abstract}
Current diffusion models for audio-driven avatar video generation struggle to synthesize long videos with natural audio synchronization and identity consistency. This paper presents StableAvatar, the first end-to-end video diffusion transformer that synthesizes infinite-length high-quality videos without post-processing. Conditioned on a reference image and audio, StableAvatar integrates tailored training and inference modules to enable infinite-length video generation. 
We observe that the main reason preventing existing models from generating long videos lies in their audio modeling. They typically rely on third-party off-the-shelf extractors to obtain audio embeddings, which are then directly injected into the diffusion model via cross-attention. Since current diffusion backbones lack any audio-related priors, this approach causes severe latent distribution error accumulation across video clips, leading the latent distribution of subsequent segments to drift away from the optimal distribution gradually.
To address this, StableAvatar introduces a novel Time-step-aware Audio Adapter that prevents error accumulation via time-step-aware modulation. During inference, we propose a novel Audio Native Guidance Mechanism to further enhance the audio synchronization by leveraging the diffusion’s own evolving joint audio-latent prediction as a dynamic guidance signal. To enhance the smoothness of the infinite-length videos, we introduce a Dynamic Weighted Sliding-window Strategy that fuses latent over time. Experiments on benchmarks show the effectiveness of StableAvatar both qualitatively and quantitatively. 

\end{abstract}    

\begin{figure}[t!]
\begin{center}
\includegraphics[width=1\linewidth]{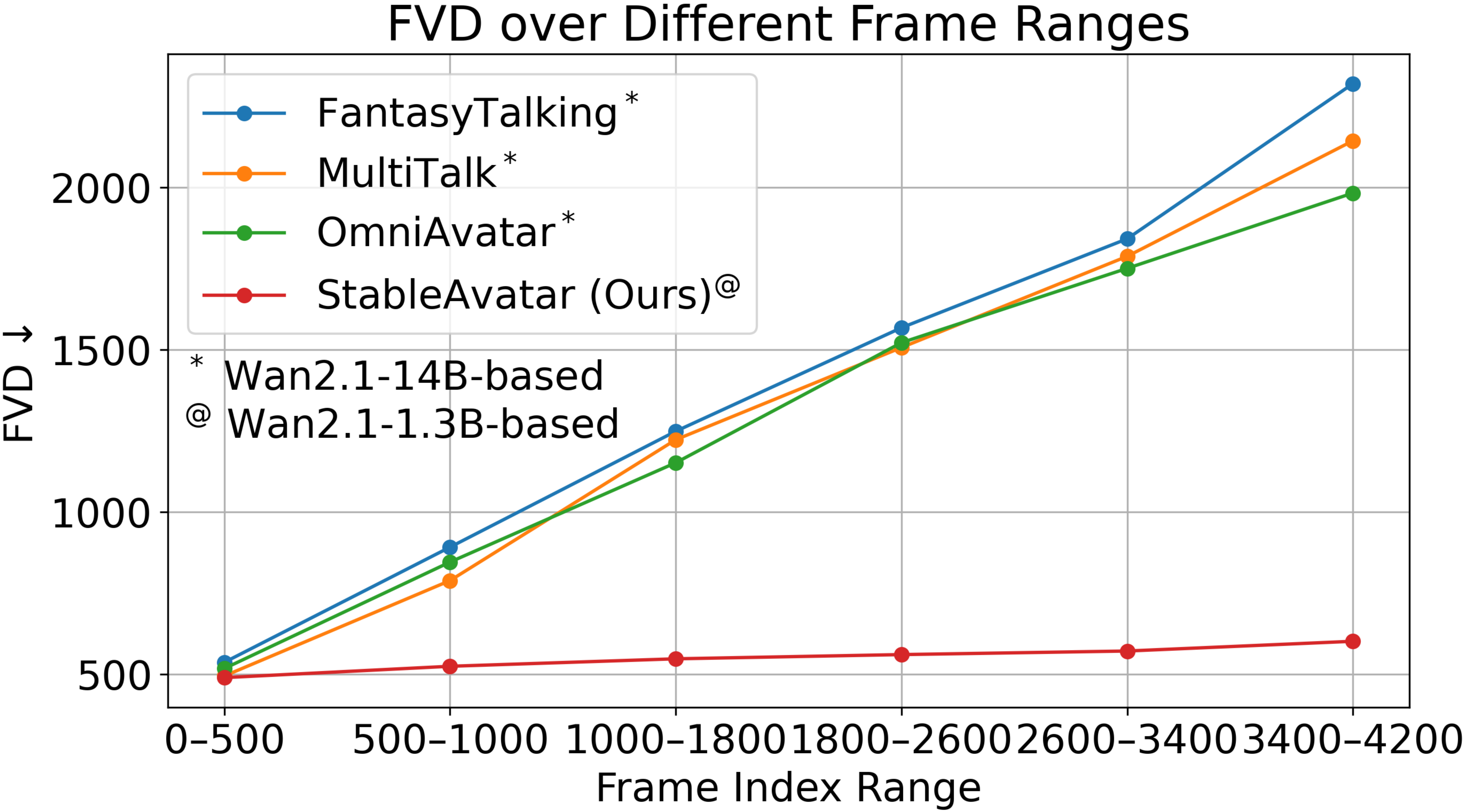}
\end{center}
\vspace{-0.6cm}
   \caption{Quantitative comparisons in quality drift.
   a-b on the x-axis is the frame range from the a-th frame to the b-th frame.
   }
\label{fig:cover_curve}
\vspace{-0.55cm}
\end{figure}

\vspace{-0.6cm}

\section{Introduction}
\label{sec:intro}
Diffusion models~\cite{xing2024simda,xing2024survey,xing2024aid,dhariwal2021diffusion,ho2020denoising,ho2022cascaded,song2020score,song2020denoising,rombach2022high,meng2021sdedit,hertz2022prompt,tumanyan2023plug,weng2024genrec,tu2024motioneditor,tu2024motionfollower,tu2025stableanimator,tu2025stableanimator++,li2025magicmotion} have significantly inspired research in audio-driven avatar video generation~\cite{zhang2023sadtalker, wei2024aniportrait, wang2024v_express, meng2025echomimicv2, cui2025hallo3, wang2025fantasytalking, chen2025hunyuanvideo, kong2025let, gan2025omniavatar, ji2025sonic}. In particular, audio-driven avatar video generation aims to synthesize natural human-centric videos with synchronized facial expressions and body movements based on a reference image and audio, offering diverse applications in film production and virtual assistants. However, current approaches are limited to generating short avatar videos of less than 15 seconds, and when they attempt to generate videos longer than 15 seconds, significant body distortions and appearance inconsistencies occur, particularly in facial regions. This severely restricts their practical applications.

To address this issue, several methods have explored consistency preservation for avatar video generation~\cite{ji2025sonic, cui2025hallo3, kong2025let, gan2025omniavatar}, yet limited effort has been devoted to tackling the underlying essence of the problem. These strategies—whether leveraging motion frames~\cite{xu2024hallo} or adopting various shifted-window mechanisms during inference—only partially improve the smoothness of long videos and fail to fundamentally mitigate quality degradation in infinite-length avatar videos.
One could alternatively split a long audio into segments, process each independently, and then concatenate them to form a continuous video. However, this approach introduces inconsistencies and abrupt transitions between segments. Thus, for audio-driven avatar video generation, end-to-end synthesis of infinite-length avatar videos while ensuring high fidelity remains an extremely challenging task.

\begin{figure*}[t!]
\begin{center}
\includegraphics[width=1\linewidth]{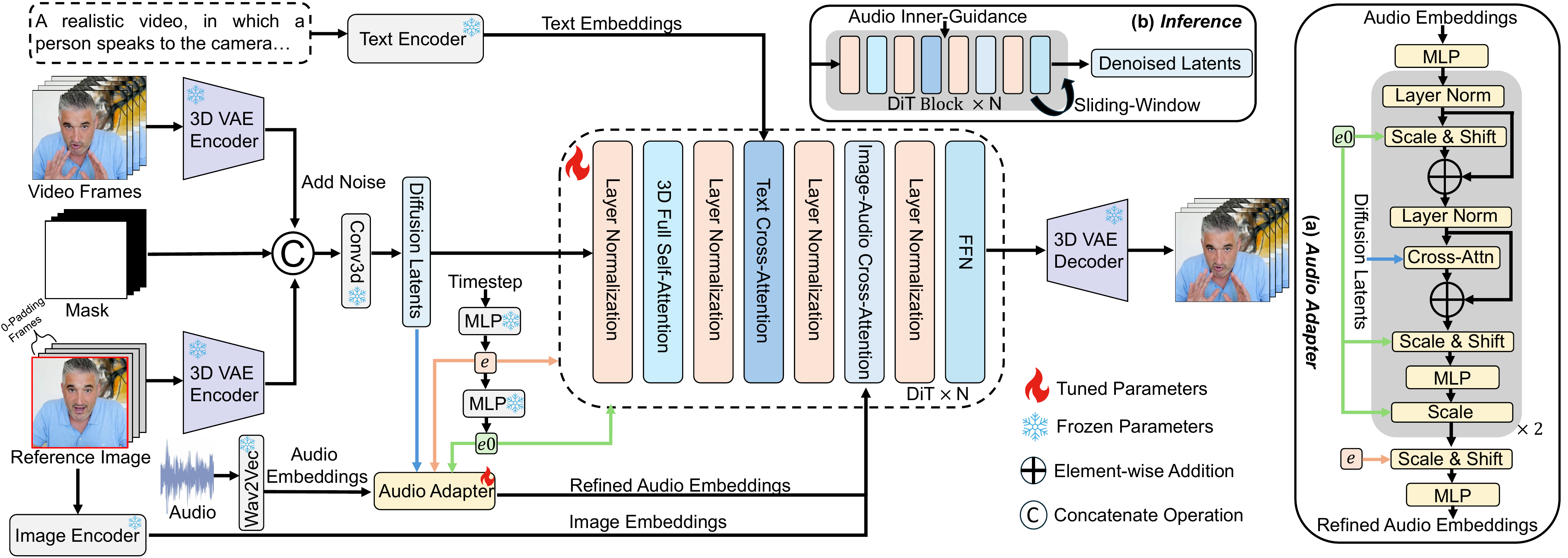}
\end{center}
\vspace{-0.6cm}
   \caption{Architecture of StableAvatar. (a) refers to the structure of the Audio Adapter. Embeddings from the Image Encoder and Text Encoder are injected to each block of DiT. Given the audio, we extract the audio embeddings utilizing Wav2Vec. 
   To model joint audio-latent representations, the audio embeddings are fed into the Audio Adapter, and its outputs are injected into the DiT via cross-attention.
   }
\label{fig:framework}
\vspace{-0.5cm}
\end{figure*}

In light of this, we propose StableAvatar, consisting of dedicated modules for both training and inference to maintain ID consistency and audio synchronization for audio-driven high-quality and infinite-length avatar video generation, as shown in Fig. \ref{fig:framework}. 
We first observe that the primary limitation preventing previous models from synthesizing infinite-length videos lies in their audio modeling. They simply use a third-party off-the-shelf extractor~\cite{baevski2020wav2vec} to obtain audio embeddings, and then directly inject them into a Video Diffusion Transformer via cross-attention. As current diffusion backbones lack any audio-related priors, this approach results in severe latent distribution error accumulation across video clips, leading the latent distribution of subsequent segments to gradually drift away from the target distribution.
To tackle this, StableAvatar feeds audio embeddings to a Video Diffusion Transformer with a novel Timestep-aware Audio Adapter that dramatically reduces error accumulation over clips.
In particular, initial audio embeddings (Query) sequentially cross-attend with initial patchified latents (Key and Value), followed by affine modulation with timestep embeddings to obtain the refined audio embeddings. 
As timestep embeddings are strongly correlated with latents, this potentially forces the diffusion to model the joint audio-latent feature distribution at each timestep, effectively mitigating latent distribution error accumulation due to the lack of audio priors.
Following~\cite{wang2025fantasytalking, kong2025let, gan2025omniavatar}, the refined audio embeddings (Key and Value) are injected into the diffusion model via cross-attention with latents (Query).

During inference, to further enhance the audio synchronization and facial expression, StableAvatar introduces a novel Audio Native Guidance Mechanism to replace the conventional Classify-Free-Guidance (CFG)~\cite{ho2022classifier}. As the refined audio embeddings are also inherently dependent on the latents, rather than solely relying on external audio signals, our guidance directly manipulates the diffusion’s sampling distribution, steering the generation towards the joint audio-latent distribution and enabling the diffusion model to refine its outputs throughout the denoising process.
StableAvatar further proposes a dynamic weighted sliding-window denoising strategy that fuses latents over time to improve the smoothness in long avatar video generation.

As shown in Fig. \ref{fig:cover} and Fig. \ref{fig:cover_curve}, while the latest audio-driven avatar video generation model OmniAvatar~\cite{gan2025omniavatar} suffers from dramatic face/body distortion and color drift, StableAvatar accurately animates the reference with natural lip synchronization driven by audio while preserving the overall consistency, even in the infinite-length video generation scenario (3500+ frames in a single pass).

In conclusion, our contributions are as follows:
(1) We propose a novel Timestep-aware Audio Adapter to force the diffusion model to capture joint audio-latent features, thereby significantly reducing latent distribution error accumulation during audio injection, making StableAvatar the first video diffusion transformer that generates infinite-length audio-driven avatar video end-to-end.
(2) A novel training-free Audio Native Guidance Mechanism replaces the conventional CFG to further enhance audio synchronization, while a dynamic weighted sliding-window denoising strategy improves the smoothness of synthesized long videos.
(4) Experimental results on benchmark datasets show the superiority of StableAvatar over the SOTA. Notably, based on a Wan2.1-1.3B model \cite{wan2025}, we surpass previous Wan2.1-14B-based models in video quality for long-video generation.
\section{Related Work}
\label{sec:related_work}

\noindent\textbf{Video Generation.}
The capacity for diversity and high fidelity in diffusion models~\cite{dhariwal2021diffusion,ho2020denoising,ho2022cascaded, tu2023implicit,nichol2021improved,song2020score,song2020denoising,rombach2022high,meng2021sdedit,hertz2022prompt,tumanyan2023plug} has sparked significant interest in their potential for video generation. 
Early video diffusion models~\cite{singer2022make, blattmann2023stable, guo2023animatediff, wu2023tune, wang2024magicvideo, tu2024motioneditor, videoworldsimulators2024,tu2024motionfollower,tu2025stableanimator} primarily leverage the U-Net architecture for video generation by adding temporal layers to pre-trained image diffusion models for joint spatio-temporal modeling.
Recent works~\cite{bao2024vidu, hong2022cogvideo, kong2024hunyuanvideo, wan2025} replace the U-Net with the Diffusion-in-Transformer (DiT) architecture~\cite{peebles2023scalable} for stronger scalability.  Inspired by previous works~\cite{wang2025fantasytalking, kong2025let, gan2025omniavatar}, we utilize Wan2.1~\cite{wan2025} as the backbone.

\noindent\textbf{Audio-Driven Avatar Video Generation.}
Audio-driven avatar video generation aims to animate a reference human image based on given audio.
Early works~\cite{guan2023stylesync, zhang2023sadtalker, cheng2022videoretalking, pang2023dpe, yin2022styleheat, gong2023toontalker, wang2024v_express} first convert audio into features~\cite{tran2018nonlinear, song2022audio}, and then use a motion-to-video model (GAN~\cite{goodfellow2020generative}) projects these features into videos. 
Recently, some studies have applied diffusion models to this field. Most works~\cite{tian2024emo, wei2024aniportrait, ji2025sonic, chen2025echomimic, li2024latentsync, jiang2024loopy} leverage the diffusion model to integrate audio cues with lips, which only supports head movement. 
CyberHost~\cite{lin2025cyberhost} and FantasyTalking~\cite{wang2025fantasytalking} aim to animate half-body avatars. 
EMO2~\cite{tian2025emo2} and EchomimicV2~\cite{meng2025echomimicv2} use hand pose sequences to improve half-body video quality. 
OmniHuman~\cite{lin2025omnihuman} applies a mixed data training strategy for scaling up.
HunyuanVideo-Avatar~\cite{chen2025hunyuanvideo} and MultiTalk~\cite{kong2025let} aim to synthesize multi-character animation. OmniAvatar~\cite{gan2025omniavatar} supports full-body avatar video generation.
However, previous works suffer from face/body distortion and color drift, particularly when generating long avatar videos. While some works~\cite{cui2025hallo3, ji2025sonic, kong2025let, gan2025omniavatar, xu2024hallo} propose strategies for long video generation, their approaches are only activated during inference and primarily improve smoothness, without addressing the underlying cause of distortion—latent error accumulation.
StableAvatar addresses these issues and performs infinite-length high-quality avatar video generation.

\noindent\textbf{Long Video Generation Strategy.}
Generating long videos~\cite{skorokhodov2022stylegan, brooks2022generating, harvey2022flexible, voleti2022mcvd, tseng2023consistent, yin2023nuwa, lu2024freelong, lu2025freelong++} is extremely challenging due to temporal complexity and the need for content consistency. 
Nuwa-XL~\cite{yin2023nuwa} and StreamingT2V~\cite{henschel2025streamingt2v} both adopt short-long memory blocks to ensure long video consistency. Gen-L-Video~\cite{wang2023gen} and FreeNoise~\cite{qiu2023freenoise} extend videos by merging overlapping sub-segments with a sliding window. FreeLong~\cite{lu2024freelong} and FreeLong++~\cite{lu2025freelong++} blend global and local video features for consistency.
However, audio-driven avatar video generation cannot directly apply the above methods, as they are designed for Text-to-Video tasks and exhibit a significant domain gap with audio.
\section{Method}
\label{sec:method}
Illustrated in Fig. \ref{fig:framework}, StableAvatar is based on the commonly used Wan2.1~\cite{wan2025} following previous works~\cite{wang2025fantasytalking, kong2025let, gan2025omniavatar}.
The audio is first fed to Wav2Vec~\cite{baevski2020wav2vec} to gain audio embeddings, which are subsequently refined for reducing latent distribution error accumulation via our Audio Adapter.
The refined audio embeddings are then fed to the denoising DiT.
More details are described in Sec. \ref{sec: audio_adapter}.
Following~\cite{wan2025}, a reference image is processed through the diffusion model via two pathways: 
(1) Concatenated with zero-filled frames along the temporal axis and transformed into a latent code by a frozen 3D VAE Encoder~\cite{wan2025}. The latent code is further concatenated with compressed video frames and binary masks (the first frame is 1 and all subsequent frames are 0) in the channel axis.
(2) Encoded by the CLIP Image Encoder~\cite{radford2021learning} to obtain image embeddings, which are fed to each image-audio cross-attention block of a denoising DiT, to modulate the synthesized appearance. 

We replace the original input video frames with random noise during inference, while the other inputs stay the same. 
We propose a novel Audio Native Guidance to replace conventional CFG~\cite{ho2022classifier} for further facilitating lip synchronization and facial expression, as detailed in Sec. \ref{sec: inner_guidance}.
We introduce a dynamic weighted sliding-window denoising strategy that fuses latent over time to enhance the video smoothness during long video generation, as detailed in Sec. \ref{sec: sliding_window}.

\subsection{Timestep-aware Audio Adapter}
\label{sec: audio_adapter}
Our goal is to generate infinite-length avatar videos under the guidance of the audio while preserving the content consistency.
Previous works~\cite{cui2025hallo3, wang2025fantasytalking, gan2025omniavatar, chen2025hunyuanvideo, kong2025let, ji2025sonic, meng2025echomimicv2} exhibit significant face and body distortions, along with color drift, in avatar videos longer than 15 seconds. This issue is attributed to their audio modeling, which directly injects the third-party off-the-shelf audio embeddings~\cite{baevski2020wav2vec} into diffusion models via cross-attention.
Current diffusion backbones~\cite{blattmann2023stable, wan2025, kong2024hunyuanvideo} lack audio-related priors, causing significant latent distribution errors to accumulate across video clips when injecting audio embeddings into the diffusion. This results in the latent distribution of subsequent segments gradually drifting away from the optimal distribution. To address this, we propose a novel Timestep-aware Audio Adapter, in which the audio embeddings go through multiple affine modulations and cross-attention blocks to interact with timestep embeddings and latents as shown in Fig. \ref{fig:framework}.

Concretely, following~\cite{tian2024emo, kong2025let}, we first use Wav2Vec~\cite{baevski2020wav2vec} to extract the raw audio embeddings $\bm{a}$. As the current state is influenced by preceding and succeeding audio frames, we then concatenate them proximal to the current frames, gaining audio embeddings $\bm{emb}_{aud}$:
\begin{equation}\small
\label{eq:audio_embeddings}
\begin{aligned}
     \bm{emb}_{aud}^{i}=\mathtt{Concat}(\bm{a}^{i-k},..., \bm{a}^{i},...,\bm{a}^{i+k}),
\end{aligned}
\end{equation}
where $2k+1$ is the context length. We further feed $\bm{emb}_{aud}$ to our Audio Adapter for addressing the error accumulation issue. Given a timestep, following~\cite{wan2025}, the DiT uses two successive MLP layers to obtain overall timestep embeddings $\bm{e}\in{\cal{R}}^{1 \times D}$ and projected timestep embeddings $\bm{e0}\in{\cal{R}}^{6 \times D}$. $D$ refers to the latent dimension. 
In diffusion pretraining, latents are tightly coupled with the timestep embeddings. Each timestep embedding corresponds to a distinct latent distribution, revealing a strong correlation between the latents and the timestep embeddings.
Due to this strong correlation, applying timestep-aware affine modulations to $\bm{emb}_{aud}$ can implicitly bridge the joint relationship between $\bm{emb}_{aud}$ and latents $\bm{z}_{i}$, enabling the diffusion model to more effectively capture joint audio-latent features, thereby overcoming the scarcity of audio priors. The modulations (scale and shift) are the same as those in the DiT for domain consistency:
\begin{equation}\small
\label{eq:audio_embeddings}
\begin{aligned}
     \bm{emb}_{aud}^{\lambda}&=\mathtt{LN}(\mathtt{MLP}(\bm{emb}_{aud}), \\
     \bm{emb}_{aud}^{\gamma}&=\bm{e0}[2]*(\bm{emb}_{aud}^{\lambda}*(1+\bm{e0}[1])+\bm{e0}[0])+\bm{emb}_{aud}^{\lambda},
\end{aligned}
\end{equation}
where $\mathtt{LN}(\cdot)$ is Layer Norm. $\mathtt{MLP}(\cdot)$ aims to project $\bm{emb}_{aud}$ onto the latent dimension. To further explicitly enhance joint audio-latent modeling, we perform cross-attention between $\bm{emb}_{aud}^{\gamma}$ (Query) and $\bm{z}_{t}$ (Key and Value), with the outputs modulated by $\bm{e0}$ as follows:
\begin{equation}\small
\label{eq:modulation}
\begin{aligned}
     \bm{emb}_{aud}^{'}&=\mathtt{CAttn}(\mathtt{LN}(\bm{emb}_{aud}^{\gamma}), \bm{z}_{t})+\mathtt{LN}(\bm{emb}_{aud}^{\gamma}), \\
     \bm{emb}_{aud}^{\eta}&=\mathtt{MLP}(\mathtt{LN}(\bm{emb}_{aud}^{'})*(1+\bm{e0}[4])+\bm{e0}[3]), \\
     \bm{emb}_{aud}^{*}&=\bm{emb}_{aud}^{'}+\bm{e0}[5]*\bm{emb}_{aud}^{\eta},
\end{aligned}
\end{equation}
where $\mathtt{CAttn}(\bm{x}, \bm{y})$ refers to cross-attention operation, in which $\bm{x}$ is the Query and $\bm{y}$ is the Value and Key. 
To more comprehensively establish the joint relationship between latents and audio representations, $\bm{emb}_{aud}^{*}$ is modulated by $\bm{e}$ to gain the refined audio embeddings $\bar{\bm{a}}_{t}$:
\begin{equation}\small
\label{eq:final_modulation}
\begin{aligned}
     \bm{e}&=\mathtt{Repeat}(\bm{e})+\bm{r}, \\
     \bar{\bm{a}}_{t}&=\mathtt{MLP}(\bm{emb}_{aud}^{*}*(1+\bm{e}[1])+\bm{e}[0]),
\end{aligned}
\end{equation}
where $\mathtt{Repeat}(\cdot)$ and $\bm{r}$ refer to duplicating along the channel dimension twice and learnable parameters.
We ultimately inject $\bar{\bm{a}}_{t}$ into the DiT via cross attention:
\begin{equation}\small
\label{eq:injection}
\begin{aligned}
     \bar{\bm{z}}_{t}&=\mathtt{CAttn}(\bm{z}_{t}, \bar{\bm{a}}_{t})+\mathtt{CAttn}(\bm{z}_{t},\bm{emb}_{img}),
\end{aligned}
\end{equation}
where $\bm{emb}_{img}$ refers to the image embeddings.

\subsection{Audio Native Guidance}
\label{sec: inner_guidance}
To further enhance audio synchronization and facial expression, we propose a novel Audio Native Guidance mechanism to replace the conventional CFG~\cite{ho2022classifier}, which does not take joint audio-latent relationship into consideration.
We modify the denoising score function~\cite{song2020score} to steer the denoising process forward in a way that maximizes both audio synchronization and naturalness.
In particular, according to our Audio Adapter, $\bar{\bm{a}}_{t}$ depends on the latents and the given audio. Thus, we treat $\bar{\bm{a}}_{t}$ as an additional prediction of the DiT, guiding the diffusion model to capture the joint audio-latent distribution, conditioned on the external signal and model parameters. The denoising process is given by:
\begin{equation}\small 
\label{eq:sampling}
\begin{aligned}
     \bar{\bm{p}}_{\theta}([\bm{z}_{t}, \bar{\bm{a}}_{t}] | \bm{A})\propto\bm{p}_{\theta}([\bm{z}_{t}, \bar{\bm{a}}_{t}] | \bm{A})\bm{p}_{\theta}(\bm{A}|[\bm{z}_{t}, \bar{\bm{a}}_{t}])^{\alpha}\bm{p}_{\theta}(\bar{\bm{a}}_{t}|\bm{z}_{t}, \bm{A})^{\beta},
\end{aligned}
\end{equation}
where $\bar{\bm{p}}_{\theta}(\cdot)$, $\bm{p}_{\theta}(\cdot)$, and $\bm{A}$ refer to the modified sampling process, original sampling process, and audio. 
$\alpha$ and $\beta$ are guidance scales.
Following Bayes’ Theorem, we obtain:
\begin{equation}\small 
\label{eq:bayes}
\begin{aligned}
     &\bm{p}_{\theta}([\bm{z}_{t}, \bar{\bm{a}}_{t}] | \bm{A})(\frac{\bm{p}_{\theta}([\bm{z}_{t}, \bar{\bm{a}}_{t}],\bm{A})}{\bm{p}_{\theta}(\bm{z}_{t}, \bar{\bm{a}}_{t})})^{\alpha}(\frac{\bm{p}_{\theta}([\bm{z}_{t}, \bar{\bm{a}}_{t}],\bm{A})}{\bm{p}_{\theta}(\bm{z}_{t}, \bm{A})})^{\beta} \\
     \rightarrow&\quad \bm{p}_{\theta}([\bm{z}_{t}, \bar{\bm{a}}_{t}] | \bm{A})(\frac{\bm{p}_{\theta}([\bm{z}_{t}, \bar{\bm{a}}_{t}]|\bm{A})\bm{p}_{\theta}(\bm{A})}{\bm{p}_{\theta}(\bm{z}_{t}, \bar{\bm{a}}_{t})})^{\alpha}(\frac{\bm{p}_{\theta}([\bm{z}_{t}, \bar{\bm{a}}_{t}]|\bm{A})}{\bm{p}_{\theta}(\bm{z}_{t}|\bm{A})})^{\beta}.
\end{aligned}
\end{equation}
As $\bm{p}_{\theta}(\bm{A})$ is a constant probability, we remove it as follows:
\begin{equation}\small 
\label{eq:constant}
\begin{aligned}
     \bm{p}_{\theta}([\bm{z}_{t}, \bar{\bm{a}}_{t}] | \bm{A})(\frac{\bm{p}_{\theta}([\bm{z}_{t}, \bar{\bm{a}}_{t}]|\bm{A})}{\bm{p}_{\theta}(\bm{z}_{t}, \bar{\bm{a}}_{t})})^{\alpha}(\frac{\bm{p}_{\theta}([\bm{z}_{t}, \bar{\bm{a}}_{t}]|\bm{A})}{\bm{p}_{\theta}(\bm{z}_{t}|\bm{A})})^{\beta}.
\end{aligned}
\end{equation}
We further convert Eq. \ref{eq:constant} to the score function format:
\begin{equation}\small 
\label{eq:score_function}
\begin{aligned}
     &(1+\alpha+\beta)\nabla_{\theta}\log\bm{p}_{\theta}([\bm{z}_{t}, \bar{\bm{a}}_{t}]|\bm{A}) \\
     &-\alpha\nabla_{\theta}\log\bm{p}_{\theta}([\bm{z}_{t}, \bar{\bm{a}}_{t}])-\beta\nabla_{\theta}\log\bm{p}_{\theta}(\bm{z}_{t}|\bm{A}).
\end{aligned}
\end{equation}
Thus, the inference formulation can be described as:
\begin{equation}\small 
\label{eq:inference}
\begin{aligned}
     &(1+\alpha+\beta)\bm{D}([\bm{z}_{t}, \bar{\bm{a}}_{t}],\bm{y}, \bm{I}, \bm{A};\theta) \\
     &-\alpha\bm{D}([\bm{z}_{t}, \bar{\bm{a}}_{t}],\bm{y}, \bm{I}, \emptyset;\theta)-\beta\bm{D}([\bm{z}_{t}, \emptyset], \bm{y}, \bm{I}, \bm{A};\theta),
\end{aligned}
\end{equation}
where $\bm{D}(\cdot)$, $\bm{y}$, and $\bm{I}$ refer to the diffusion model, text prompt, and reference image. Notably, $\bm{I}$ and $\bm{y}$ are not guidance factors, as we find that incorporating $\bm{I}$ and $\bm{y}$ into the guidance substantially increases GPU resource consumption and does not significantly improve visual quality.
The Audio Native Guidance mechanism regards $\bar{\bm{a}}_{t}$ as an additional prediction target for the diffusion model, allowing the model to be guided by the joint audio-latent distribution, thereby ensuring that audio and latents are strongly correlated during the denoising process. It significantly reduces distribution error accumulation in audio-driven video generation, even when the base model lacks audio priors.

\begin{algorithm}[t!]
\caption{Dynamic Weighted Sliding-Window Strategy}
\label{alg:sliding_window}
\begin{algorithmic}[1]
\small 
\State \textbf{Input:} {$\bm{emb}_{aud}^{[0,L]}$, $T$, $\bm{z}_{T}^{[0,L]}$, $\bm{D}(\cdot)$, $l~(l<L)$, $m$, $\bm{y}$, $\bm{I}$} 
    \State \textbf{for} $t \in \{T, \ldots, 1\}$ \textbf{do} \hfill $\triangleright$ $T$ is denoising steps
        \State \hspace{0.5em} $\text{start index }s=0$ \hfill $\triangleright$ $m$ is overlap length between windows
        \State \hspace{0.5em} $\text{end index }e=s+l$ \hfill $\triangleright$ $l$ is video clip/window length
        \State \hspace{0.5em} $\text{previous end index }e_{prev}=e$ \hfill $\triangleright$ $\bm{z}_{T}^{[0,L]}$ are noised latents
        \State \hspace{0.5em} \textbf{while} $e\le L$ \textbf{do} \hfill $\triangleright$ $\bm{I}$ is the reference image
        \State \hspace{1.2em} $\bm{z}_{t-1}^{[s, e]}=\bm{D}(\bm{z}_{t}^{[s, e]},\bm{emb}_{aud}^{[s, e]},t,\bm{y},\bm{I})$ \hfill $\triangleright$ $\bm{y}$ is the text
        \State \hspace{1.2em} \textbf{if} $s\ne0$ and $t \ne T$: \hfill $\triangleright$ $\mathtt{log1p}(x)$ is $\log(x+1)$
        \State \hspace{1.8em} $\bm{w}=\mathtt{np.linspace}(0,1,\text{num\_samples=}m)$
        \State \hspace{1.8em} $\bm{w}=\mathtt{np.log1p}(\bm{w} * (\mathtt{np.exp}(1) - 1))$
        \State \hspace{1.8em} $\bm{w}=\frac{\bm{w}-\mathtt{min}(\bm{w})}{\mathtt{max}(\bm{w})-\mathtt{min}(\bm{w})}$
        \State \hspace{1.8em} $\bm{z}_{t-1}^{[s,s+m]}=\bm{w}*\bm{z}_{t-1}^{[s,s+m]}+(1-\bm{w})*\bm{z}_{t-1}^{[e_{prev}-m,e_{prev}]}$
        \State \hspace{1.2em}  \textbf{if} $e<L$: \hfill $\triangleright$ \text{It's for the edge case of the final clip}
        \State \hspace{1.8em} $e_{prev}=e, s=s+(l-m)$
        \State \hspace{1.8em}  $e=\mathtt{min}(s+l, L)$
        \State \hspace{1.2em}  \textbf{else}: \textbf{break}
    \State \textbf{return} $\bm{z}_{0}^{[0,L]}$
\end{algorithmic}
\vspace{-0.1cm}
\end{algorithm}

\subsection{Dynamic Weighted Sliding-Window Strategy}
\label{sec: sliding_window}
To improve the smoothness of synthesized long avatar videos, we further propose a Dynamic Weighted Sliding-Window Strategy (DWSW) during inference.
Compared with the previous sling-window denoising strategy~\cite{ji2025sonic},  overlapping latents between adjacent windows are fused using a sliding window mechanism, where the fusion weights follow a logarithmic interpolation based on the relative frame indices, as described in the Algorithm \ref{alg:sliding_window} and Fig. \ref{fig:sliding_window}.
$L$ is the VAE-compressed total video length.
The fused latents are injected back into both adjacent windows, ensuring that both boundaries of the central window consist of blended features. Leveraging a logarithmic weighting function introduces a progressive smoothing effect in the transitions between video clips. Early stages experience a more pronounced weight variation, while later stages exhibit subtle shifts, resulting in seamless continuity across video clips.

\subsection{Training}
\label{sec: training}
We use the reconstruction loss to train our model, with trainable components including attention modules of a Dit and an Audio Adapter.
We introduce face masks $\bm{M}_{face}$ and lip masks $\bm{M}_{lip}$, extracted by Mediapipe~\cite{lugaresi2019mediapipe} from the input video frames to enhance the modeling of face regions:
\begin{equation}\small
\label{eq:loss}
\begin{aligned}
     \mathcal{L}=\begin{cases}
  \mathbb{E}_{\theta}(\left \| (\bm{z}_{gt}-\bm{z}_{\theta})\odot  \bm{M}_{face}  \right \|^{2}) ,~~~0.4\le\bm{q}<0.5& \\
  \mathbb{E}_{\theta}(\left \| (\bm{z}_{gt}-\bm{z}_{\theta})\odot  \bm{M}_{lip}  \right \|^{2}) ,~~~~~~~\bm{q}\ge0.5&  \\
  \mathbb{E}_{\theta}(\left \| (\bm{z}_{gt}-\bm{z}_{\theta})\odot  (1+\bm{M}_{face}+\bm{M}_{lip})  \right \|^{2}) , \text{otherwise}& 
\end{cases}
\end{aligned}
\end{equation}
where $\bm{z}_{gt}$ and $\bm{z}_{\varepsilon}$ refer to diffusion latents and denoised latents, respectively. $\bm{q}$ is a random variable uniformly distributed over the interval $[0, 1]$.
This piecewise objective separately supervises lip synchronization and facial expression, enabling more targeted learning.
\section{Experiments}
\label{sec:experiments}

\subsection{Implementation Details}
Our training dataset is composed of three parts: Hallo3~\cite{cui2025hallo3}, Celebv-HQ~\cite{zhu2022celebvhq}, and videos collected from the internet, totaling 1200 hours of video.
Following previous works~\cite{wang2025fantasytalking,cui2025hallo3,kong2025let,gan2025omniavatar}, we evaluate our model on HDTF~\cite{zhang2021flow} and semi-body AVSpeech~\cite{ephrat2018looking}. 
Since previous works do not open-source their testing datasets, we randomly select 100 videos (5-20 seconds long) from HDTF and AVSpeech, respectively.
We conduct additional experiments on 100 unseen videos (2-5 minutes long), referred to the Long100, selected from the internet to assess the robustness of our model during long avatar animation.
Our DiT uses pre-trained weights of Wan2.1-I2V 1.3B~\cite{wan2025, liu2025phantom}, while the Audio Encoder is trained from scratch. 
Our model is trained for 20 epochs on 64 NVIDIA A100 80G GPUs, with a batch size of 1 per GPU. 
We set learning rate=1$e$-5, $\alpha=4.5$, and $\beta=3.0$.

\begin{table*}[t!]\small
\caption{Quantitative comparisons on HDTF/AVSpeech/Long100. In the table elements $a$ / $b$ / $c$, $a$, $b$, and $c$ refer to the result on the HDTF, AVSpeech, and Long100, respectively. The average video duration of HDTF and AVSpeech is 10 seconds, while Long100 is 3 minutes.
}
\vspace{-0.25in}
\begin{center}
\renewcommand\arraystretch{1.1}
\scalebox{0.85}{
\begin{tabular}{l|ccccccc}
\toprule
Model          & FID$\downarrow$                                    & FVD$\downarrow$                              & CSIM$\uparrow$                                  & Sync-C$\uparrow$                             & Sync-D$\downarrow$                             & IQA$\uparrow$                                & ASE$\uparrow$                                \\ \midrule
SadTalker~\cite{zhang2023sadtalker}      & 53.12/120.57/194.88                    & 567/1468/2261                    & 0.821/0.802/0.423                     & 7.25/4.13/3.21                     & 9.86/9.72/11.16                    & 3.12/2.40/2.31                     & 2.14/1.43/1.38                     \\
Aniportrait~\cite{wei2024aniportrait}    & 46.35/118.86/190.66                    & 537/1524/2095                    & 0.815/0.796/0.415                     & 3.88/1.95/1.13                     & 10.84/11.58/12.87                  & 3.82/2.28/2.10                     & 2.41/1.31/1.22                     \\
Sonic~\cite{ji2025sonic}          & 62.17/187.42/278.40                    & 552/2051/2877                    & 0.843/0.825/0.451                     & 8.16/6.04/4.03                     & 7.65/8.94/10.23                    & 3.28/2.24/2.08                     & 2.05/1.36/1.13                     \\
EchoMimic~\cite{chen2025echomimic}      & 63.43/108.13/178.12                    & 593/1123/1885                    & 0.849/0.837/0.456                     & 5.44/4.59/3.64                     & 9.27/9.78/10.74                    & 3.64/2.97/2.31                     & 2.23/1.82/1.43                     \\
Hallo3~\cite{cui2025hallo3}         & 44.31/98.14/170.44                     & 438/987/1724                     & 0.845/0.834/0.462                     & 6.58/5.16/4.42                     & 8.64/9.62/9.92                     & 3.50/3.38/2.35                     & 2.11/1.96/1.36                     \\
FantasyTalking~\cite{wang2025fantasytalking} & 46.74/80.01/175.78                     & 479/823/1789                     & 0.863/0.853/0.468                     & 3.42/2.98/1.92                     & 12.15/11.422/11.78                 & 3.55/3.24/2.42                     & 2.28/1.89/1.48                     \\
HunyuanAvatar~\cite{chen2025hunyuanvideo}  & 52.16/77.53/172.94                     & 625/868/1743                     & 0.866/0.859/0.472                     & 7.20/6.74/4.34                     & 8.39/8.28/10.07                    & 3.56/3.63/2.46                     & 2.25/2.21/1.52                     \\
MultiTalk~\cite{kong2025let}      & 46.94/75.67/175.52                     & 446/804/1768                     & 0.868/0.861/0.465                     & 7.53/4.88/4.12                     & 8.02/9.59/10.18                    & 3.54/3.72/2.42                     & 2.15/2.25/1.45                     \\
OmniAvatar~\cite{gan2025omniavatar}     & 41.79/72.56/168.49         & 424/744/1621         & 0.862/0.857/0.471          & 7.50/6.78/4.45          & 8.26/8.05/9.62          & 3.55/3.74/2.51          & 2.27/2.29/1.56 \\ \midrule
Ours           & \textbf{38.14/68.12/57.18}             & \textbf{375/640/504}             & \textbf{0.875/0.872/0.849}            & \textbf{8.15/7.56/8.24}            & \textbf{6.94/7.85/6.79}            & \textbf{3.90/3.79/3.84}            & \textbf{2.46/2.32/2.39}            \\ \bottomrule
\end{tabular}
}
\end{center}
\vspace{-0.3in}
\label{table:quantitative_comparisons}
\end{table*}

\subsection{Comparison with State-of-the-Art Methods}
\textbf{Quantitative results.}
Regarding metrics, we use FID~\cite{heusel2017gans} and FVD~\cite{unterthiner2018towards} to assess the quality of synthesized images and videos.
We further use the Q-align model~\cite{wu2023q} to evaluate the video quality (IQA) and aesthetic metrics (ASE). Sync-C~\cite{chung2016out} and Sync-D~\cite{chung2016out} are utilized to assess the synchronization of lips with audio.
CSIM~\cite{zhang2023sadtalker} evaluates the cosine similarity between the facial embeddings of two images. 
We compare with recent audio-driven avatar video generation models, including GAN-based models (SadTalker~\cite{zhang2023sadtalker}) and diffusion-based models (SD-based: AniPortrait~\cite{wei2024aniportrait}, EchoMimic~\cite{chen2025echomimic}; SVD-based: Sonic~\cite{ji2025sonic}; CogVideo-5B-based: Hallo3~\cite{cui2025hallo3}; HunyuanVideo-13B-based: HunyuanAvatar~\cite{chen2025hunyuanvideo}; Wan-14B-based: FantasyTalking~\cite{wang2025fantasytalking}, MultiTalk~\cite{kong2025let}, OmniAvatar~\cite{gan2025omniavatar}).
Based on previous studies that assess quantitative results using the self-driven and reconstruction approach, we perform quantitative comparisons with the above competitors on HDTF~\cite{zhang2021flow}, AVSpeech~\cite{ephrat2018looking}, and Long100.
Notably, all competitors are trained on our dataset before evaluating on Long100 to ensure a fair comparison.
The results are shown in Table \ref{table:quantitative_comparisons}.
We observe that, even though all competitors encounter a significant drop in performance for long video generation, StableAvatar still surpasses them regarding face quality, video fidelity, and lip synchronization while maintaining relatively high single-frame quality.
Specifically, Wan2.1-1.3B-based StableAvatar outperforms the leading competitor, Wan2.1-14B-based OmniAvatar, by 80.3\% and 85.2\% in CSIM and Sync-C on Long100.

\begin{figure*}[t!]
\begin{center}
\includegraphics[width=0.98\linewidth]{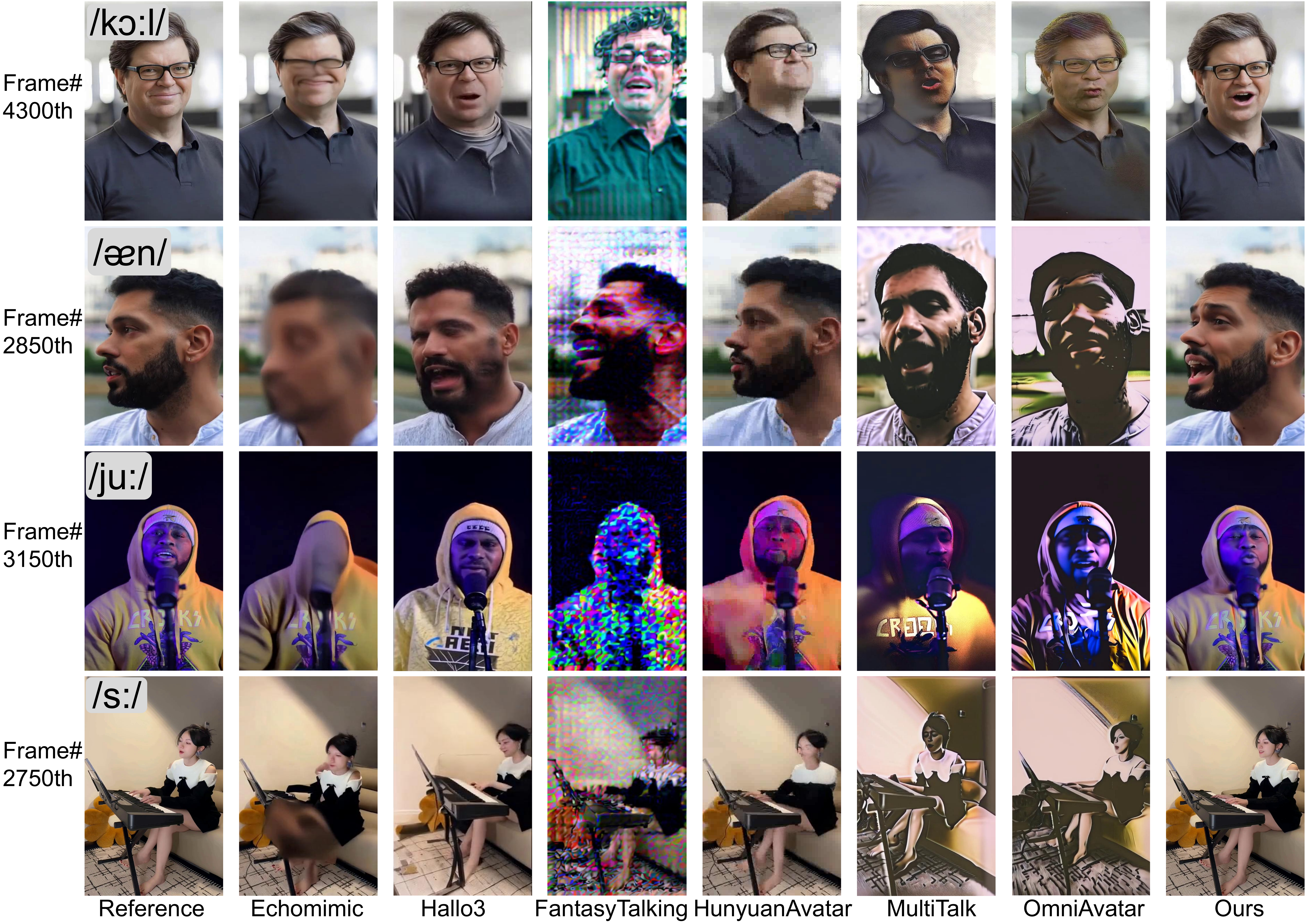}
\end{center}
\vspace{-0.65cm}
   \caption{Qualitative comparisons with state-of-the-art methods. More examples can be found in the supplementary material.}
\label{fig:comparison}
\vspace{-0.5cm}
\end{figure*}

\noindent\textbf{Qualitative Results.}
The qualitative results are shown in Fig. \ref{fig:comparison}. Notably, each audio lasts 3+ minutes and is filled with intricate rhythm patterns, while the references include intricate details of appearances. We only display selected frames from the last 2 minutes for brevity.
EchoMimic~\cite{chen2025echomimic} exhibits face/body distortion and clothing changes, while the rest of competitors can accurately modify the reference lip movements within the first 15 seconds of the video.
However, when the video duration exceeds 15 seconds, all competitors suffer from audio-lip synchronization issues, blurry noises, face distortion, and color drift.
In particular, Hallo3~\cite{cui2025hallo3} and HunyuanAvatar~\cite{chen2025hunyuanvideo} suffer from severe face distortion and audio-lip synchronization issues, with the lips moving randomly. Meanwhile, FantasyTalking~\cite{wang2025fantasytalking}, MultiTalk~\cite{kong2025let}, and OmniAvatar~\cite{gan2025omniavatar} struggle with color drift, body/face distortion, and audio-lip synchronization issues.
In contrast, our StableAvatar accurately animates images based on the given audio while preserving reference identities even after generating 3500+ frames, highlighting the superiority of our model in identity retention and in generating vivid, infinite-length avatar videos.

\noindent\textbf{Length Discussion.}
Fig. \ref{fig:cover_curve} shows that the quality drift in our StableAvatar remains negligible as the frame count increases, especially when compared to previous models. Theoretically, our StableAvatar is capable of synthesizing hours of video without significant quality degradation.

\begin{figure}[t!]
\begin{center}
\includegraphics[width=1\linewidth]{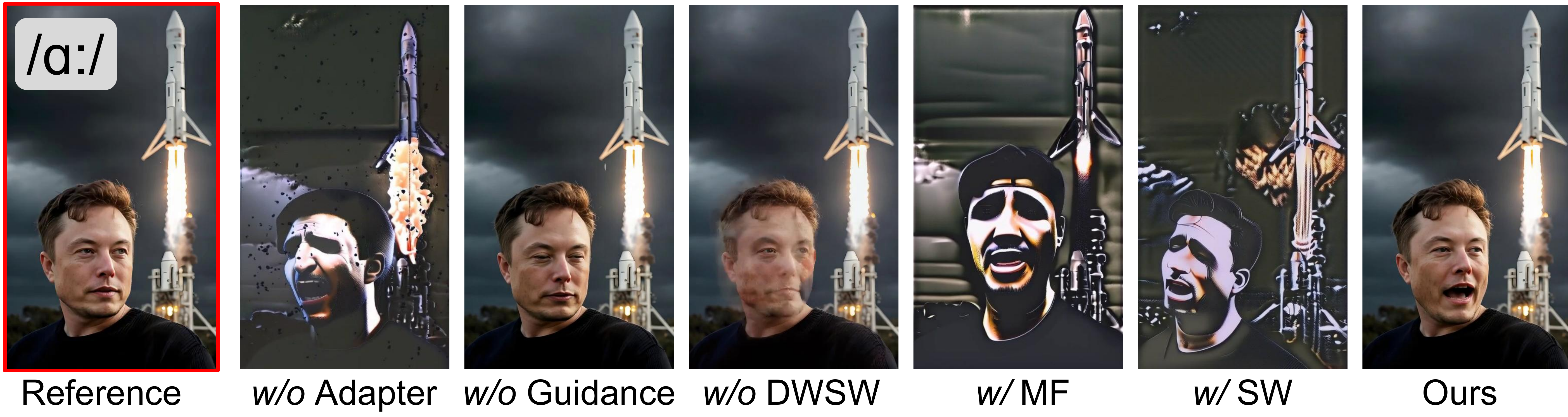}
\end{center}
\vspace{-0.7cm}
   \caption{Ablations on core components of StableAvatar. 
   }
\label{fig:ablation_core}
\vspace{-0.40cm}
\end{figure}

\begin{table}[t!]\small
\caption{Ablation study on core components. MF and SW are motion frame~\cite{cui2025hallo3} and conventional sliding-window~\cite{ji2025sonic}, which are both based on \textit{w/o} Audio Adapter.
}
\vspace{-0.25in}
\begin{center}
\renewcommand\arraystretch{1.1}
\scalebox{0.75}{
\begin{tabular}{l|cccccc}
\toprule
Method            & FVD$\downarrow$          & CSIM$\uparrow$           & Sync-C$\uparrow$        & Sync-D$\downarrow$        & IQA$\uparrow$           & ASE$\uparrow$           \\ \midrule
\textit{w/o} Audio Adapter & 1802         & 0.457          & 3.95          & 10.96         & 2.34          & 1.40          \\
\textit{w/o} Guidance      & 866          & 0.822          & 7.48          & 8.36          & 3.74          & 2.31          \\
\textit{w/o} DWSW          & 718          & 0.845          & 8.17          & 6.85          & 3.79          & 2.36          \\
\textit{w/} MF~\cite{cui2025hallo3}   & 2043         & 0.402          & 3.69          & 10.82         & 2.28          & 1.32          \\
\textit{w/} SW~\cite{ji2025sonic} & 1854         & 0.438          & 3.77          & 10.72         & 2.31          & 1.35          \\ \midrule
Ours              & \textbf{504} & \textbf{0.849} & \textbf{8.24} & \textbf{6.79} & \textbf{3.84} & \textbf{2.39} \\ \bottomrule
\end{tabular}
}
\end{center}
\label{table:ablation_core}
\vspace{-0.3in}
\end{table}

\subsection{Ablation Study}

\textbf{Audio Adapter.}
We conduct an ablation study to demonstrate the contributions of core components in StableAvatar, as shown in Table \ref{table:ablation_core} and Fig. \ref{fig:ablation_core}. Notably, all quantitative ablation studies are on the Long100 dataset. We can see that removing the core components significantly degrades performance, particularly in CSIM and Sync-C/D, highlighting that our components significantly enhance both video fidelity while preserving high ID consistency in long avatar video generation. In contrast, previous long video generation strategies (\textit{w/} MF~\cite{cui2025hallo3} and \textit{w/} SW~\cite{ji2025sonic}) still suffer from dramatic appearance inconsistency and color drift, as they only basically tackle the video smoothness issue.

\begin{figure}[t!]
\begin{center}
\includegraphics[width=1\linewidth]{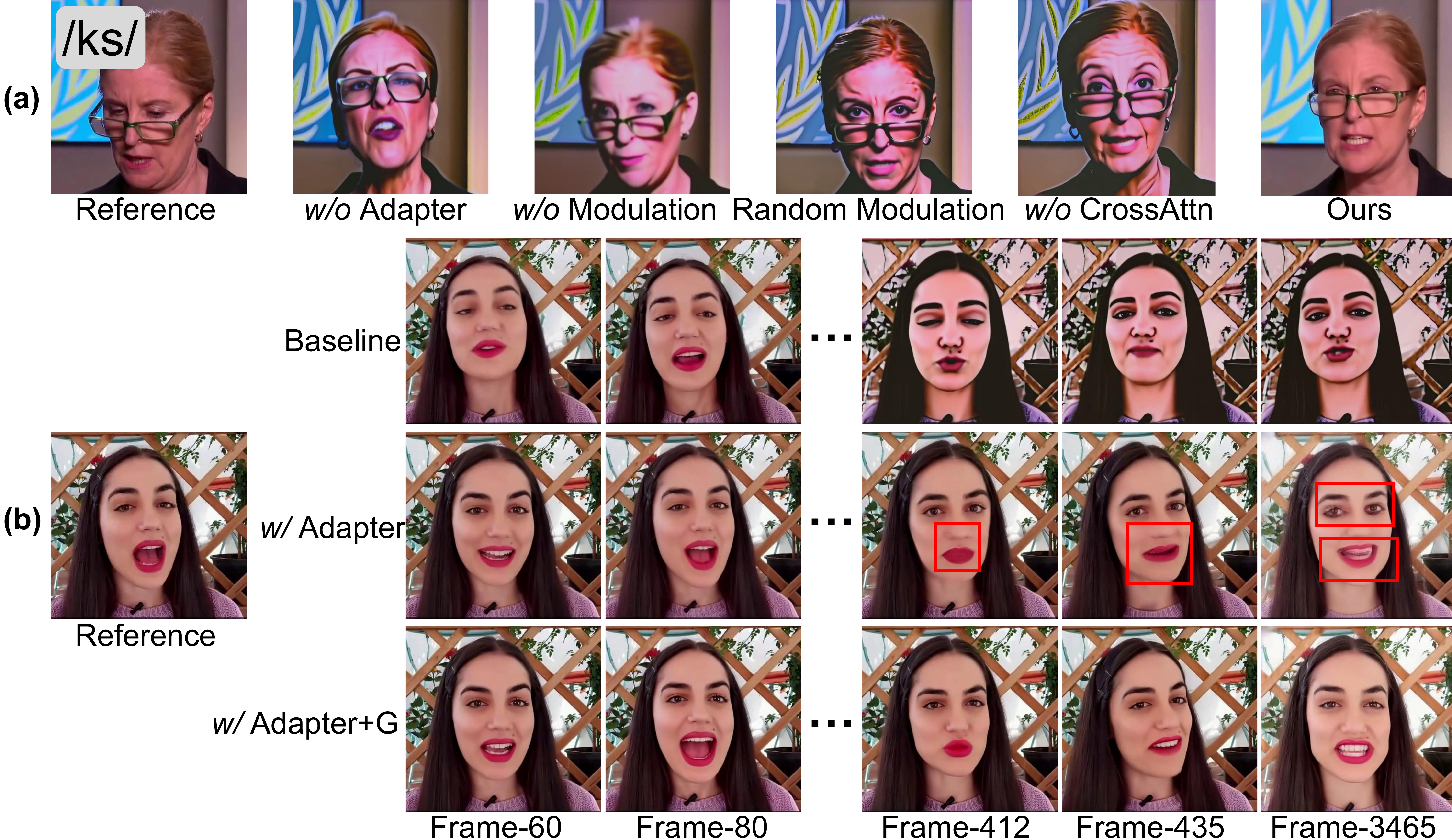}
\end{center}
\vspace{-0.6cm}
   \caption{Ablation study on audio modeling.
   }
\label{fig:ablation_audio_modeling}
\vspace{-0.45cm}
\end{figure}

We further conduct an ablation study regarding audio modeling, as shown in Table \ref{table:ablation_audio} and Fig. \ref{fig:ablation_audio_modeling} (a).
By analyzing the results, we can gain the following observations: 
(1) \textit{w/o} Aduio Adapter dramatically degrades the video fidelity and lip synchronization. The plausible reason is that current diffusion backbones lack audio-related priors, and directly injecting audio embeddings from third-party off-the-shelf extractors into diffusion leads to significant latent error accumulation across video clips, gradually deteriorating the overall long video quality.
(2) \textit{w/o} Modulation/CAttn both relatively degrade the video quality. The underlying reason is that timestep-aware modulation bridges the joint audio-latent modeling, as latents and timesteps are strongly correlated. CAttn explicitly introduces latents into audio modeling, but without timestep modulation for audio embeddings, making it challenging for the model to effectively model the joint latent-audio space.
Thus, timestep-aware modulation and CAttn are complementary, which is also evidenced by \textit{w/} Random modulation results.
(3) StableAvatar can significantly refine the face quality while maintaining high video fidelity in long video generation since our model enables joint audio-latent modeling, reducing latent distribution error accumulation across video clips.

\begin{table}[t!]\small
\caption{Ablation study on audio modeling. \textit{w/o} Random modulation and \textit{w/o} CAttn refer to replace timesteps with random learnable parameters and remove cross-attention in our Audio Adapter.
}
\vspace{-0.25in}
\begin{center}
\renewcommand\arraystretch{1.1}
\scalebox{0.75}{
\begin{tabular}{l|cccccc}
\toprule
Method               & FVD$\downarrow$          & CSIM$\uparrow$           & Sync-C$\uparrow$        & Sync-D$\downarrow$        & IQA$\uparrow$           & ASE$\uparrow$           \\ \midrule
\textit{w/o} Audio Adapter    & 1802         & 0.457          & 3.95          & 10.96         & 2.34          & 1.40          \\
\textit{w/o} Modulation       & 1340         & 0.637          & 5.25          & 9.48          & 3.39          & 1.98          \\
\textit{w/} Random modulation & 1186         & 0.632          & 5.16          & 9.76          & 3.50          & 2.07          \\
\textit{w/o} CAttn  & 1218         & 0.664          & 6.37          & 8.52          & 3.56          & 2.23          \\ \midrule
Ours                 & \textbf{504} & \textbf{0.849} & \textbf{8.24} & \textbf{6.79} & \textbf{3.84} & \textbf{2.39} \\ \bottomrule
\end{tabular}
}
\end{center}
\label{table:ablation_audio}
\vspace{-0.3in}
\end{table}

\begin{table}[t!]\small
\caption{Ablation study on the error accumulation. Adapter and G refer to our Audio Adapter and Audio Native Guidance.
}
\vspace{-0.2in}
\begin{center}
\renewcommand\arraystretch{1.1}
\scalebox{0.8}{
\begin{tabular}{l|ccccc}
\toprule
Method          & FVD$\downarrow$          & CSIM$\uparrow$           & Sync-C$\uparrow$        & Sync-D$\downarrow$        & CIEDE$\downarrow$          \\ \midrule
Baseline(A)     & 865          & 0.836          & 7.66          & 7.82          & 0.536          \\
Baseline(B)     & 2388         & 0.405          & 3.78          & 10.47         & 2.318          \\ \midrule
\textit{w/} Adapter(A)   & 723          & 0.829          & 8.23          & 6.74          & 0.196          \\
\textit{w/} Adapter(B)   & 912          & 0.818          & 7.95          & 6.92          & 0.831          \\ \midrule
\textit{w/} Adapter+G(A) & 478          & 0.846          & 8.28          & 6.65          & 0.166          \\
\textit{w/} Adapter+G(B) & 572 & 0.853 & 8.20 & 6.83 & 0.523 \\ \bottomrule
\end{tabular}
}
\end{center}
\label{table:ablation_error}
\vspace{-0.3in}
\end{table}

\noindent\textbf{Error Accumulation.}
We conduct an ablation study on error accumulation, as shown in Table \ref{table:ablation_error} and Fig. \ref{fig:ablation_audio_modeling}(b). 
A and B pick the 1st-200th and 3500th-3700th frames for evaluation, respectively. CIEDE~\cite{sharma2005ciede2000} measures the extent of color drift.
Baseline removes all our audio-related components.
We have the following observations: 
(1) Baseline suffers from significant video quality deterioration in the 3500th-3700th frames. The main reason is that raw audio embeddings have conflicts with original backbone priors, triggering latent distribution error at each clip.
As the number of generated frames increases, error accumulation becomes more severe, and the shift of the subsequent distribution from the target distribution grows larger.
(2) \textit{w/} Adapter relatively maintains video fidelity even in the later frame intervals. It indicates that our Audio Adapter helps the diffusion model to overcome the scarcity of audio priors via timestep-aware modulation, thereby tackling the error accumulation issue.
(3) Our guidance can also ensure the video quality stability in the long video generation to some extents, as it can further alleviate the latent error at each clip. 
(4) Our audio-related components ensure that even after generating 3500+ frames, the video consistency and fidelity remain stable without significant degradation, certainly addressing the accumulation issue.

\begin{figure}[t!]
\begin{center}
\includegraphics[width=1\linewidth]{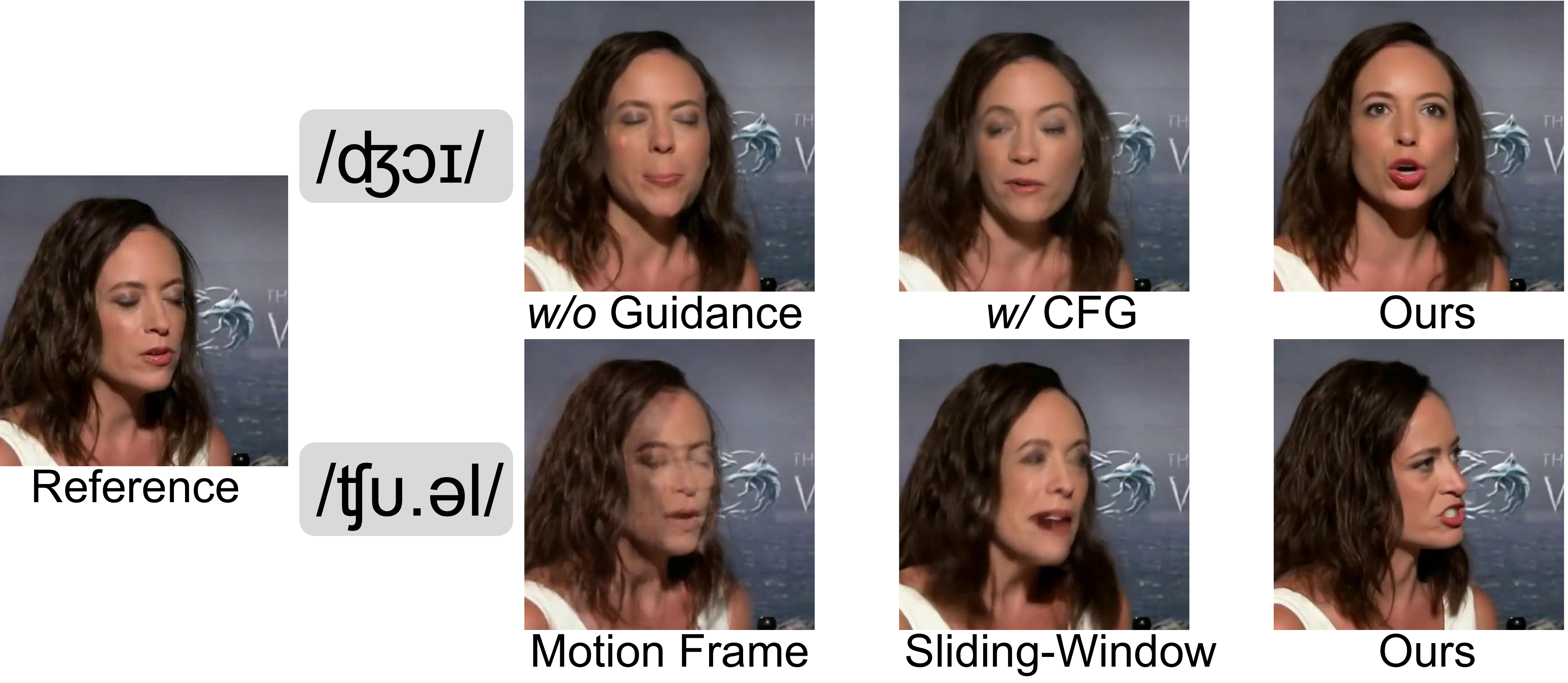}
\end{center}
\vspace{-0.8cm}
   \caption{Ablation study on long video generation strategies.
   }
\label{fig:ablation_guidance_window}
\vspace{-0.4cm}
\end{figure}

\begin{table}[t!]\small
\caption{Ablation study on the guidance. \textit{w/} CFG replaces our Audio Native Guidance with Classify-Free-Guidance (CFG).
}
\vspace{-0.25in}
\begin{center}
\renewcommand\arraystretch{1.1}
\scalebox{0.75}{
\begin{tabular}{l|cccccc}
\toprule
Method       & FVD$\downarrow$          & CSIM$\uparrow$           & Sync-C$\uparrow$        & Sync-D$\downarrow$        & IQA$\uparrow$           & ASE$\uparrow$           \\ \midrule
\textit{w/o} Guidance & 866          & 0.822          & 7.48          & 8.36          & 3.74          & 2.31          \\
\textit{w/} CFG~\cite{ho2022classifier}       & 822          & 0.828          & 7.62          & 7.91          & 3.78          & 2.33          \\ \midrule
Ours         & \textbf{532} & \textbf{0.853} & \textbf{8.20} & \textbf{6.83} & \textbf{3.84} & \textbf{2.39} \\ \bottomrule
\end{tabular}
}
\end{center}
\label{table:ablation_guidance}
\vspace{-0.3in}
\end{table}

\noindent\textbf{Audio Native Guidance.}
To validate the significance of our Audio Native Guidance, we conduct an ablation regarding different strategies. The results are in Table \ref{table:ablation_guidance} and Fig. \ref{fig:ablation_guidance_window}.
While conventional CFG only regards each external condition as an individual signal which is independent of latents, our guidance considers audio embeddings as correlated with the latents, accounting for the joint audio-latent distribution, thereby further facilitating the lip synchronization/naturalness and video fidelity. 

\begin{table}[t!]\small
\caption{Ablation study on long avatar video generation methods.
}
\vspace{-0.25in}
\begin{center}
\renewcommand\arraystretch{1.1}
\scalebox{0.75}{
\begin{tabular}{l|cccccc}
\toprule
Method         & FVD$\downarrow$          & CSIM$\uparrow$           & Sync-C$\uparrow$        & Sync-D$\downarrow$        & IQA$\uparrow$           & ASE$\uparrow$           \\ \midrule
Motion Frame~\cite{cui2025hallo3}   & 772          & 0.836          & 8.12          & 6.98          & 3.72          & 2.23          \\
Sliding Window~\cite{ji2025sonic} & 698          & 0.842          & 8.18          & 6.89          & 3.77          & 2.31          \\ \midrule
Ours           & \textbf{532} & \textbf{0.853} & \textbf{8.20} & \textbf{6.83} & \textbf{3.84} & \textbf{2.39} \\ \bottomrule
\end{tabular}
}
\end{center}
\label{table:ablation_window}
\vspace{-0.3in}
\end{table}

\noindent\textbf{Long Video Strategy.}
We conduct a comparison between our DWSW and other types of long avatar video generation strategies, as shown in Table \ref{table:ablation_window} and Fig. \ref{fig:ablation_guidance_window}.
We can see that both motion frame~\cite{cui2025hallo3, kong2025let} and conventional sliding-window~\cite{ji2025sonic} fail to eliminate the jitter caused by the connection between video clips. By contrast, our DWSW leverages a logarithmic interpolation to dynamically assign weights to different context windows, significantly mitigating the impact of video clip connection. More ablation studies are in Sec. \ref{supp:ablation_window} of the Supp.

\subsection{Applications and User Study}
\noindent\textbf{Speed and GPU Resource.}
We compare the inference speed and GPU memory consumption between our StableAvatar and previous models, as shown in Sec. \ref{supp:gpu} of the Supp.
With approximately 50\% of the memory consumption and 10 times the inference speed of the leading competitor OmniAvatar~\cite{gan2025omniavatar}, our Wan2.1-1.3B-based StableAvatar significantly outperforms previous Wan2.1-14B-based models~\cite{wang2025fantasytalking,kong2025let,gan2025omniavatar} in face quality and lip synchronization, highlighting its superiority in long avatar video generation.

\noindent\textbf{Full Body Avatar Videos.}
We conduct a qualitative experiment on our StableAvatar in full/half-body avatar animation.
The results are shown in Sec. \ref{supp:full_body} of the Supp. 
Each protagonist in the reference image interacts with an object, such as an instrument or an apple.
We can observe that our StableAvatar can handle full/half-body avatar animation in high fidelity while preserving identities even during intensive object interactions.

\noindent\textbf{Multi-Avatar Animation.}
We experiment on audio-driven multi-avatar animation, as shown in Sec. \ref{supp:multi_avatar} of the Supp. We can see that our model is capable of animating multiple individuals following the given audio. 

\noindent\textbf{Cartoon Avatars.}
To validate the robustness of our StableAvatar, we experiment on audio-driven cartoon avatar animation, as shown in Sec. \ref{supp:cartoon} of the Supp. We can observe that our model can synthesize natural cartoon avatar videos with rich facial expressions based on the given audio.

\noindent\textbf{User Study.}
We conduct a user study on 30 selected videos to evaluate the human preference between our StableAvatar and other competitors. The participants are basically university students and faculty. In each case, participants are first presented with the reference image and the audio. Then we provide two videos (one is generated by our StableAvatar and the other is synthesized by a competitor) in random orders. Participants are then asked to answer the following questions: 
L-A/A-A/B-A/I-A: ``Which one has better lip/appearance/background/ID alignment with the audio/reference". Table \ref{table:user_study} shows the superiority of our model regarding subjective evaluation.

\begin{table}[t!]\small
\caption{User preference of StableAvatar compared with other competitors. Higher indicates users prefer more to our model.
}
\vspace{-0.25in}
\begin{center}
\renewcommand\arraystretch{1.1}
\scalebox{0.9}{
\begin{tabular}{lcccc}
\toprule
Method         & L-A    & A-A    & B-A    & I-A    \\ \midrule
Hallo3~\cite{cui2025hallo3}         & 97.4\% & 98.1\% & 95.2\% & 98.9\% \\
FantasyTalking~\cite{wang2025fantasytalking} & 98.6\% & 95.8\% & 94.9\% & 98.1\% \\
HunyuanAvatar~\cite{chen2025hunyuanvideo}  & 96.2\% & 94.5\% & 94.6\% & 97.5\% \\
MultiTalk~\cite{kong2025let}      & 95.5\% & 95.2\% & 94.2\% & 96.4\% \\
OmniAvatar~\cite{gan2025omniavatar}     & 94.6\% & 95.6\% & 93.8\% & 95.8\% \\ \bottomrule
\end{tabular}
}
\end{center}
\label{table:user_study}
\vspace{-0.3in}
\end{table}
\section{Conclusion}
\label{sec:conclusion}
In this paper, we proposed StableAvatar, a video diffusion transformer with dedicated modules for training and inference to synthesize infinite-length high-quality avatar videos.
StableAvatar first utilized off-the-shelf models to gain audio embeddings. To overcome the scarcity of audio priors of diffusion backbones, StableAvatar introduced an Audio Adapter to refine audio embeddings.
In inference, to further enhance lip synchronization with audio, StableAvatar introduced an Audio Native Guidance Mechanism to replace conventional Classify-Free-Guidance. To improve the long video's smoothness, StableAvatar further proposed a dynamic weighted sliding-window strategy.
Experimental results across various datasets demonstrated the superiority of our model in producing infinite-length high-quality avatar videos.
{
    \small
    \bibliographystyle{ieeenat_fullname}
    \bibliography{main}
}

\clearpage
\setcounter{page}{1}
\appendix
\section{Supplementary Material}

\subsection{Preliminaries}
The diffusion model includes a forward diffusion process and a reverse denoising process. In the forward process, inspired by  Rectified Flow~\cite{lipman2022flow}, the Gaussian noise is progressively added to the data sample $\bm{x}_{0}\sim\bm{p}_{\text{data}}$ from the particular data distribution $\bm{p}_{\text{data}}$:
\begin{equation}\small
\label{eq:forward_diffusion}
\begin{aligned}
    \bm{x}_{t}=(1-t)\bm{x}_{0}+t\bm{e},
\end{aligned}
\end{equation}
where $t \in [0,1]$ refers to the timestep and $\bm{e}$ is sampled from a standard-normal distribution $\mathcal{N}(0,1)$.
The data sample $\bm{x}_{0}$ is ultimately converted into Gaussian noise $\bm{x}_{T}\sim\mathcal{N}(0,1)$ after $\bm{T}$ diffusion forward steps. 
In the reverse process, the diffusion model $\bm{\varepsilon}_{\theta}(\bm{x}_{t},t)$ is trained to predict the velocity $\bm{v'}_{t}=d\bm{x}_{t}/dt$ conditioned on the noisy latents $x_{t}$ and the timestep $t$. The MSE loss is applied to train $\bm{\varepsilon}(\cdot)$:
\begin{equation}\small
\label{eq:mse_loss}
\begin{aligned}
     \mathcal{L} = \mathbb{E}_{\bm{x}_{0},\bm{\varepsilon},t}(\left \| \bm{v'}_{t} -\bm{\varepsilon}_{\theta}(\bm{x}_{t}, t)  \right \|^{2}).
\end{aligned}
\end{equation}

\subsection{Sling Window Details}
\begin{figure}[t!]
\begin{center}
\includegraphics[width=1\linewidth]{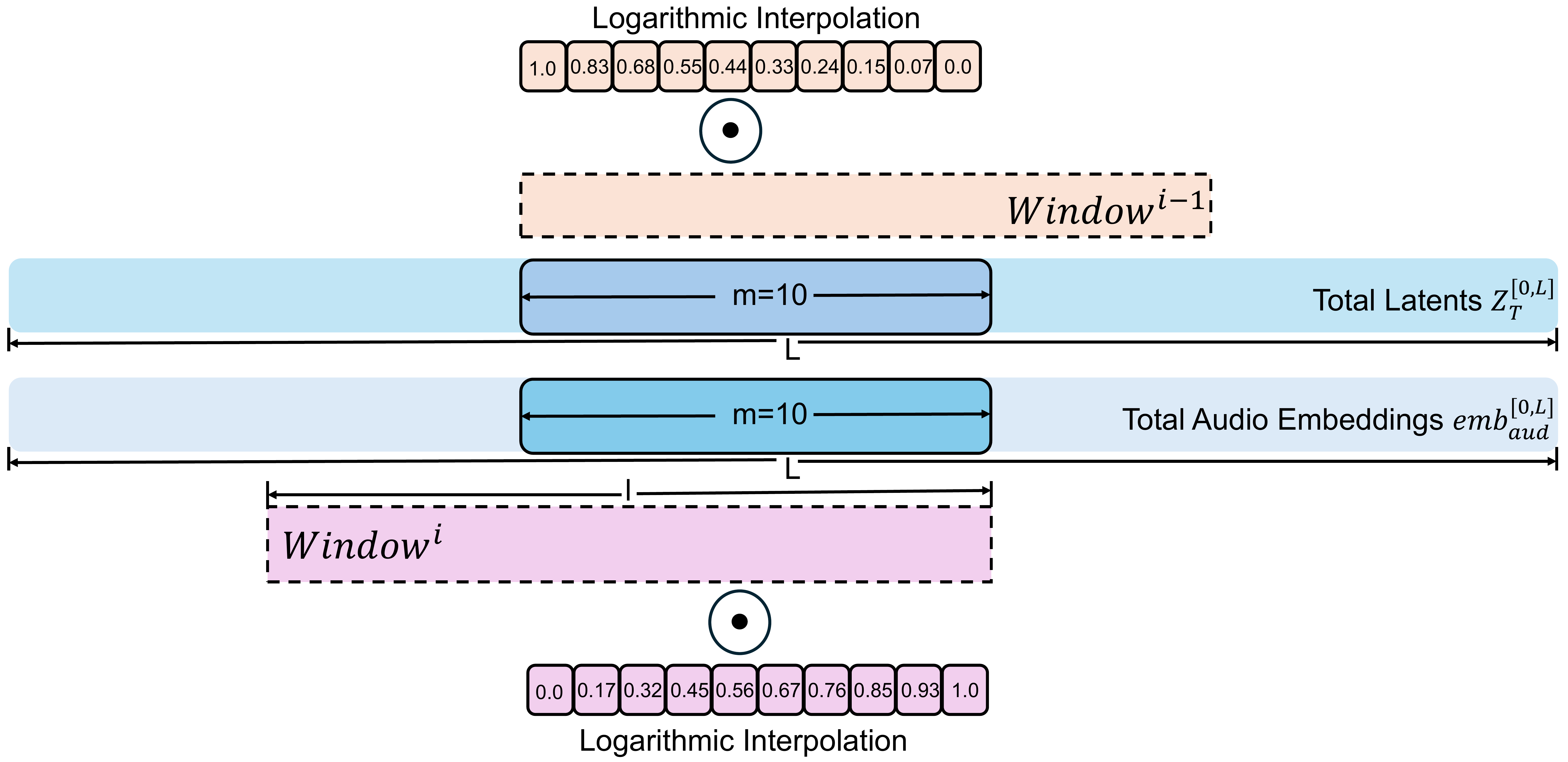}
\end{center}
\vspace{-0.5cm}
   \caption{The pipeline of our DWSW.
   }
\label{fig:sliding_window}
\vspace{-0.3cm}
\end{figure}

Fig. \ref{fig:sliding_window} illustrates a detailed pipeline of our Dynamic Weighted Sliding Window Denoising Strategy (DWSW).

\subsection{Details of Dataset}
Regarding the training dataset, our training dataset consists of three parts, including Hallo3~\cite{cui2025hallo3}, Celebv-HQ~\cite{zhu2022celebvhq}, and collected videos from the internet (BilBil, YouTube, and TikTok). 
We leverage SyncNet~\cite{chung2016out} and Q-Align~\cite{wu2023q} to filter for higher-quality videos by assessing lip-sync accuracy with audio and video fidelity.
We also employ InsightFace~\cite{deng2019arcface} to filter out videos with a facial confidence score below 0.9. Ultimately, we obtain the refined training dataset, containing approximately 1200 hours of videos.

\begin{figure}[t!]
\begin{center}
\includegraphics[width=1\linewidth]{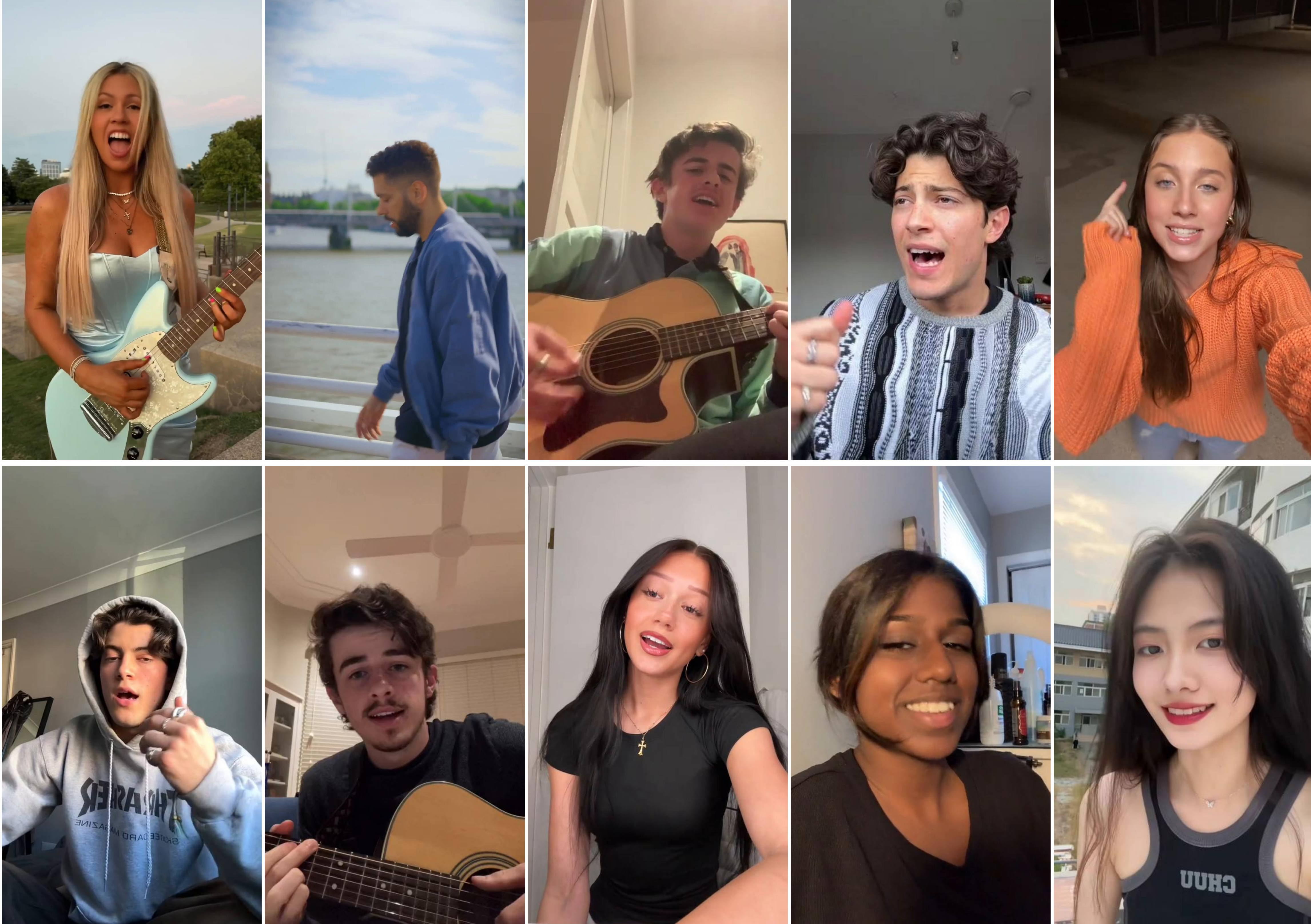}
\end{center}
\vspace{-0.5cm}
   \caption{Examples from Long100.
   }
\label{fig:long100}
\vspace{-0.3cm}
\end{figure}

In terms of the testing dataset, we select 100 unseen videos (2-5 minutes long, FPS=30) from the internet to construct the testing dataset Long100. Some examples are shown in Fig. \ref{fig:long100}.
The sources of videos come from numerous social media platforms, including YouTube, TikTok, and BiliBili. These videos showcase individuals across ethnicities, genders, portrayed in full-body, half-body, and close-up shots against varied indoor and outdoor settings.
In contrast to existing open-source testing datasets (HDTF and AVSpeech), our Long100 contains relatively complicated audio rhythms and intricate protagonist appearances. 
Moreover, the average duration of Long100 (2-5 minutes) is significantly longer than existing open-source testing datasets (2-60 seconds). Long100 also involves multiple human-object interactions, such as instruments, making it more challenging to maintain identity consistency.

\begin{table}[t!]\small
\caption{Ablation study on different weight assignment.
}
\vspace{-0.25in}
\begin{center}
\renewcommand\arraystretch{1.1}
\scalebox{0.75}{
\begin{tabular}{lcccccc}
\toprule
Model         & FVD$\downarrow$          & CSIM$\uparrow$           & Sync-C$\uparrow$        & Sync-D$\downarrow$        & IQA$\uparrow$           & ASE$\uparrow$           \\ \midrule
Fixed weights & 705          & 0.838          & 8.08          & 7.18          & 3.73          & 2.25          \\
Uniformed gap & 594          & 0.847          & 8.15          & 6.94          & 3.82          & 2.33          \\ \midrule
Ours          & \textbf{532} & \textbf{0.853} & \textbf{8.20} & \textbf{6.83} & \textbf{3.84} & \textbf{2.39} \\ \bottomrule
\end{tabular}
}
\end{center}
\label{table:weight}
\vspace{-0.2in}
\end{table}

\subsection{Additional Ablation on DWSW}
\label{supp:ablation_window}
We further conduct an ablation study on the weight assignment of our Dynamic Weighted Sliding Window Denoising Strategy (DWSM),as shown in Table \ref{table:weight}.
Fixed weights set all weights of context window to 0.5 and Uniformed gap refers to the weights of each element in the context window following an arithmetic progression.
We can observe that our logarithmic weighting function achieves the best performance, indicating that the logarithmic function prioritizes the fusion of elements from the previous context window in the early stages, while progressively and smoothly incorporating both context windows in the later stages, thereby facilitating the overall long video smoothness.

\begin{table}[t!]\small
\caption{Comparisons in inference latency and GPU resources. Inference latency (minutes) and GPU memory are measured for generating 81 frames at 480×832 resolution.
}
\vspace{-0.25in}
\begin{center}
\renewcommand\arraystretch{1.1}
\scalebox{0.75}{
\begin{tabular}{lccccc}
\toprule
Method         & CSIM$\uparrow$           & Sync-C$\uparrow$        & Sync-D$\downarrow$        & GPU Mem$\downarrow$ & Speed$\downarrow$ \\ \midrule
Hallo3~\cite{cui2025hallo3}         & 0.462          & 4.42          & 9.92          & 49.8G       & 13.72     \\
FantasyTalking~\cite{wang2025fantasytalking} & 0.468          & 1.92          & 11.78         & 44.7G       & 16.40     \\
HunyuanAvatar~\cite{chen2025hunyuanvideo}  & 0.472          & 4.34          & 10.07         & 26.6G       & 21.98     \\
MultiTalk~\cite{kong2025let}      & 0.465          & 4.12          & 10.18         & 49.21G       & 23.0     \\
OmniAvatar~\cite{gan2025omniavatar}     & 0.471          & 4.45          & 9.62          & 40.83G       & 20.31     \\ \midrule
Ours           & \textbf{0.849} & \textbf{8.24} & \textbf{6.79} & \textbf{18.4G}       & \textbf{2.32}     \\ \bottomrule
\end{tabular}
}
\end{center}
\label{table:gpu}
\vspace{-0.3in}
\end{table}

\subsection{Speed and GPU Resource.}
\label{supp:gpu}
We compare the inference speed and GPU memory consumption between our StableAvatar and previous models, as shown in Table \ref{table:gpu}.
Despite being built on a significantly smaller backbone (Wan2.1-1.3B), StableAvatar achieves superior face quality and lip synchronization compared to prior Wan2.1-14B-based models~\cite{wang2025fantasytalking,kong2025let,gan2025omniavatar}, while requiring only half the memory and achieving a tenfold increase in inference speed over the leading competitor OmniAvatar~\cite{gan2025omniavatar}. This demonstrates its strong advantage in long-form avatar video generation.

\subsection{Full/Half-Body Avatar Video Results}
\label{supp:full_body}
\begin{figure*}[t!]
\begin{center}
\includegraphics[width=1\linewidth]{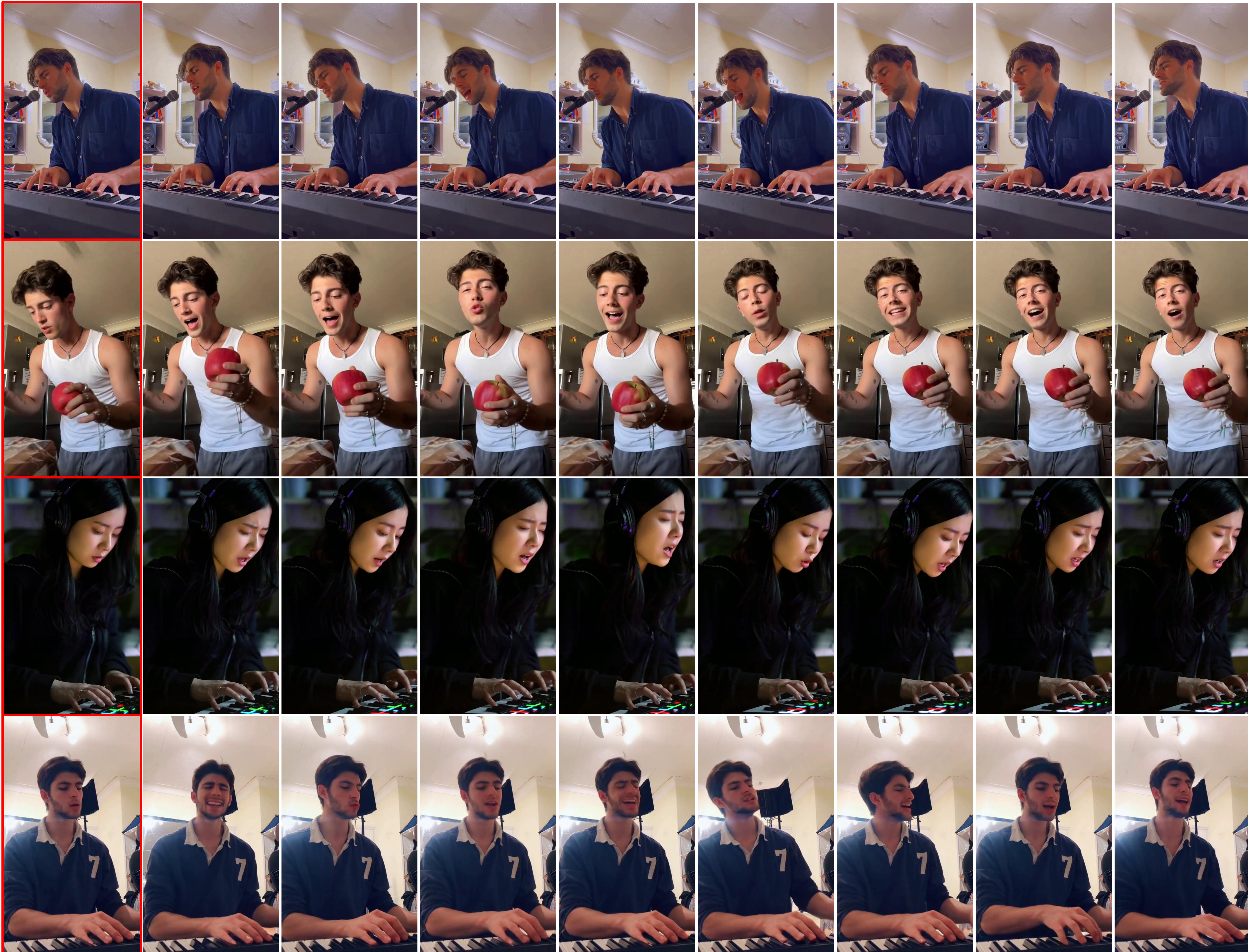}
\end{center}
\vspace{-0.55cm}
   \caption{Full/Half-body avatar animation results. The images with red borders are the reference images.}
\label{fig:full_body}
\vspace{-0.5cm}
\end{figure*}

We perform a qualitative experiment in audio-driven full/half-body avatar animations, as shown in Fig. \ref{fig:full_body}.
Each reference protagonist interacts with an object, such as an apple or a piano, making it more challenging to preserve appearance consistency and lip synchronization with the given audio.
We can see that our model is capable of synthesizing full/half-body avatar videos, even involving interactions with external objects.

\subsection{Multi-Avatar Animation}
\label{supp:multi_avatar}
\begin{figure*}[t!]
\begin{center}
\includegraphics[width=1\linewidth]{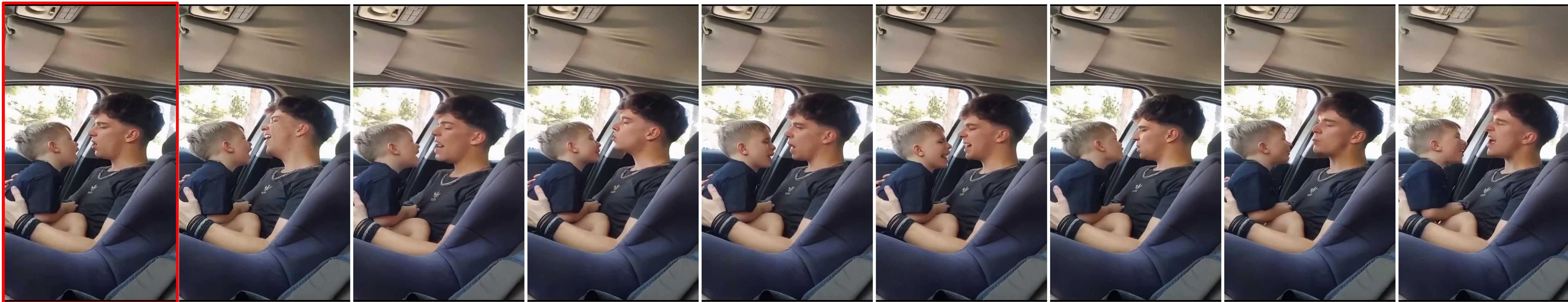}
\end{center}
\vspace{-0.55cm}
   \caption{Audio-driven multiple-avatar animation results. The images with red borders are the reference images.}
\label{fig:multiple_person}
\vspace{-0.5cm}
\end{figure*}

To demonstrate the robustness of our StableAvatar, we experiment on a particular reference image involving multiple protagonists, as shown in Fig. \ref{fig:multiple_person}. We can see that our StableAvatar is also capable of handling audio-driven multiple-avatar animations while preserving the original identity and achieving high video fidelity.

\subsection{Cartoon Avatar Video Generation}
\label{supp:cartoon}
\begin{figure*}[t!]
\begin{center}
\includegraphics[width=1\linewidth]{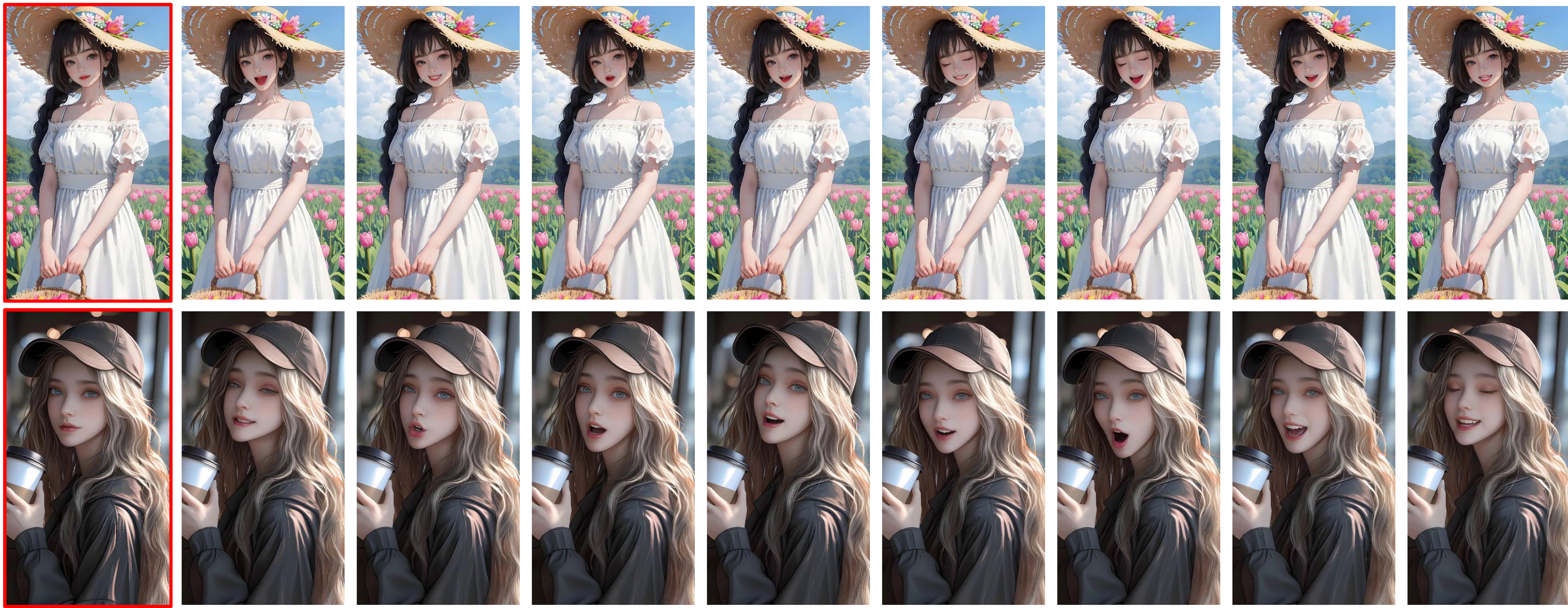}
\end{center}
\vspace{-0.55cm}
   \caption{Audio-driven cartoon avatar animation results. The images with red borders are the reference images.}
\label{fig:cartoon}
\vspace{-0.5cm}
\end{figure*}

To validate the diversity of our StableAvatar, we perform experiments on two reference images involving cartoon protagonists, as shown in Fig. \ref{fig:cartoon}. We can observe that our StableAvatar can also handle audio-driven cartoon avatar video generation, highlighting the diversity of our StableAvatar.

\subsection{Long Avatar Video Results}
\label{supp:long_video}
\begin{figure*}[t!]
\begin{center}
\includegraphics[width=1\linewidth]{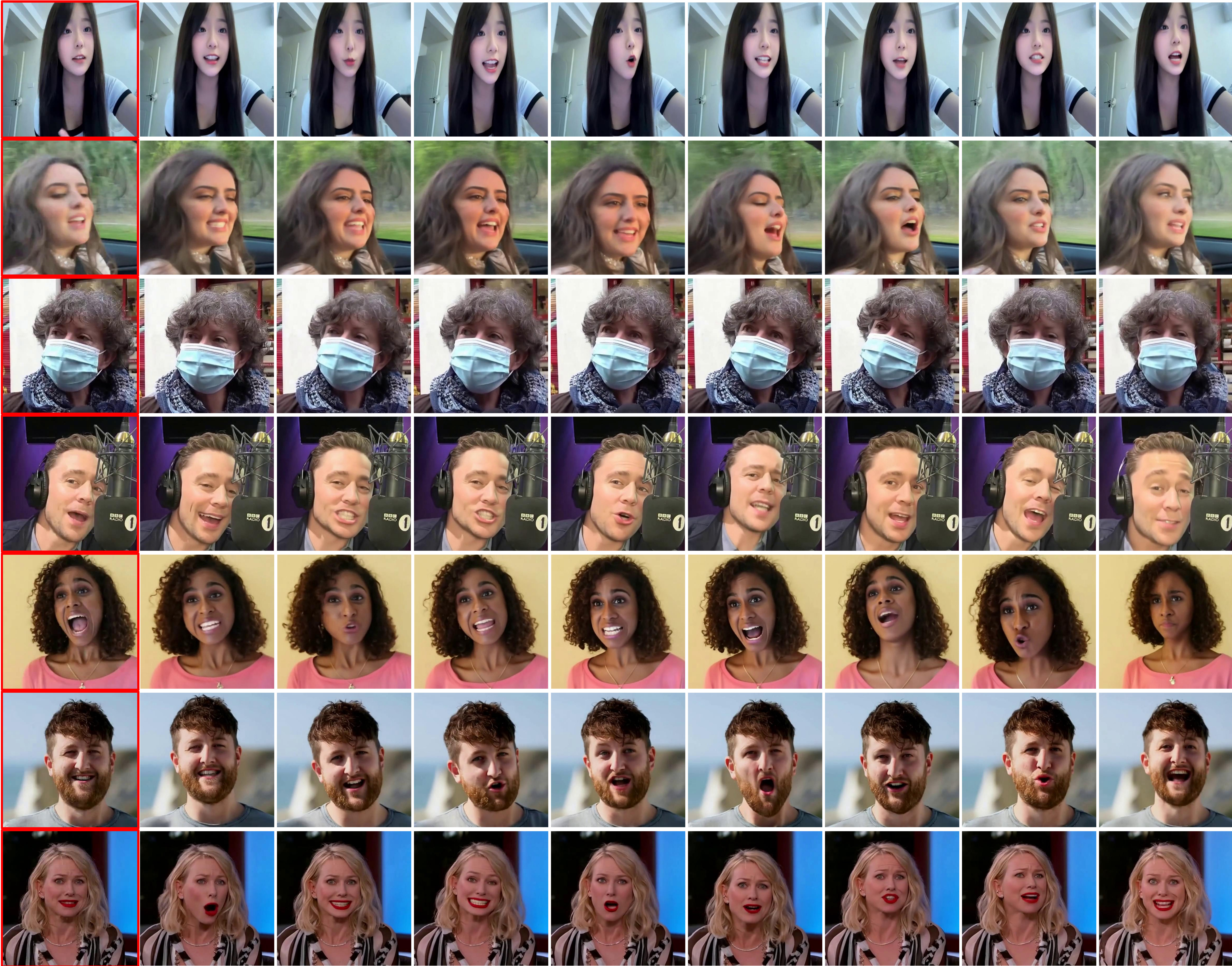}
\end{center}
\vspace{-0.55cm}
   \caption{Audio-driven long avatar video results (1/5). The images with red borders are the reference images.}
\label{fig:long_video_1}
\vspace{-0.5cm}
\end{figure*}

\begin{figure*}[t!]
\begin{center}
\includegraphics[width=1\linewidth]{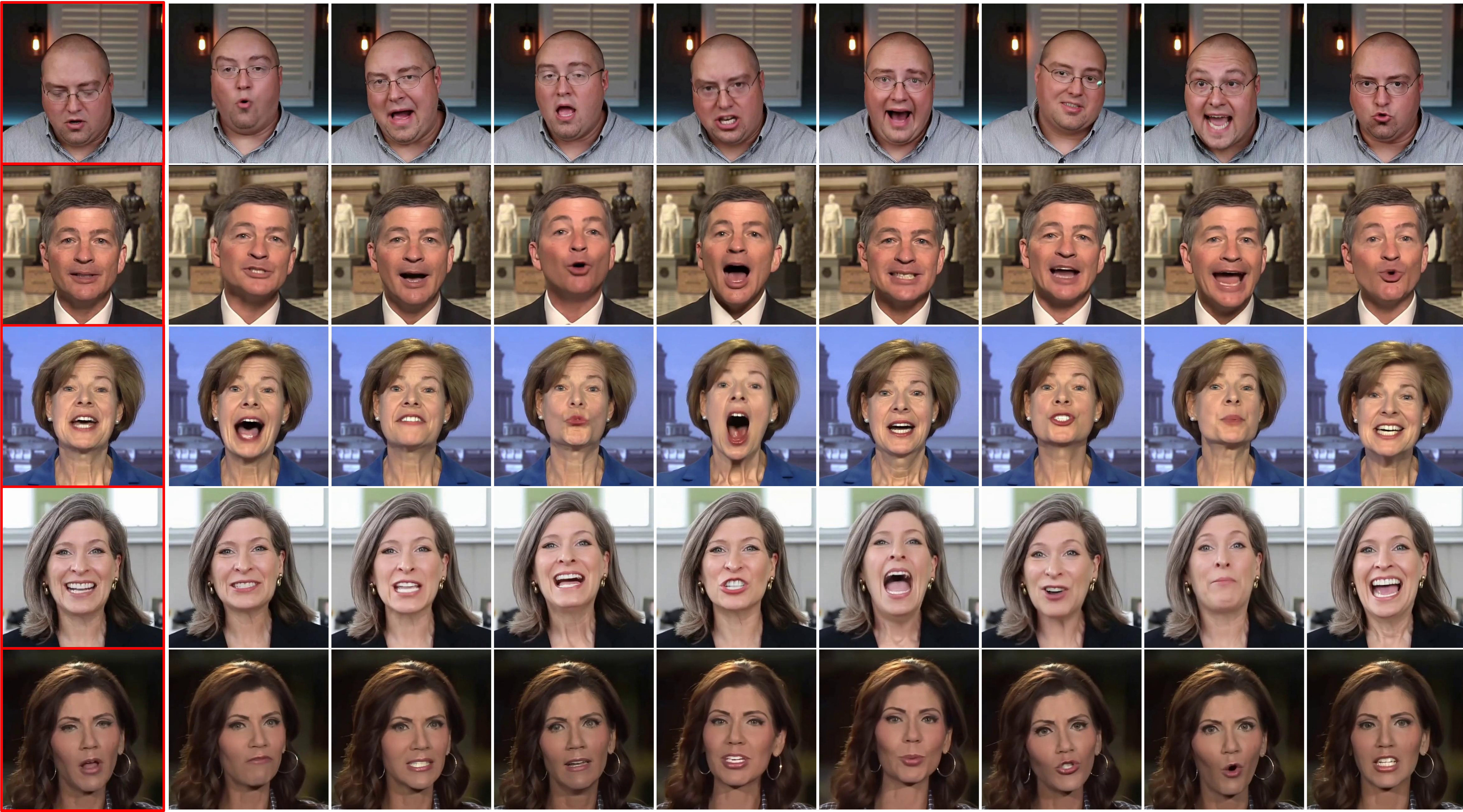}
\end{center}
\vspace{-0.55cm}
   \caption{Audio-driven long avatar video results (2/5). The images with red borders are the reference images.}
\label{fig:long_video_2}
\vspace{-0.5cm}
\end{figure*}

\begin{figure*}[t!]
\begin{center}
\includegraphics[width=1\linewidth]{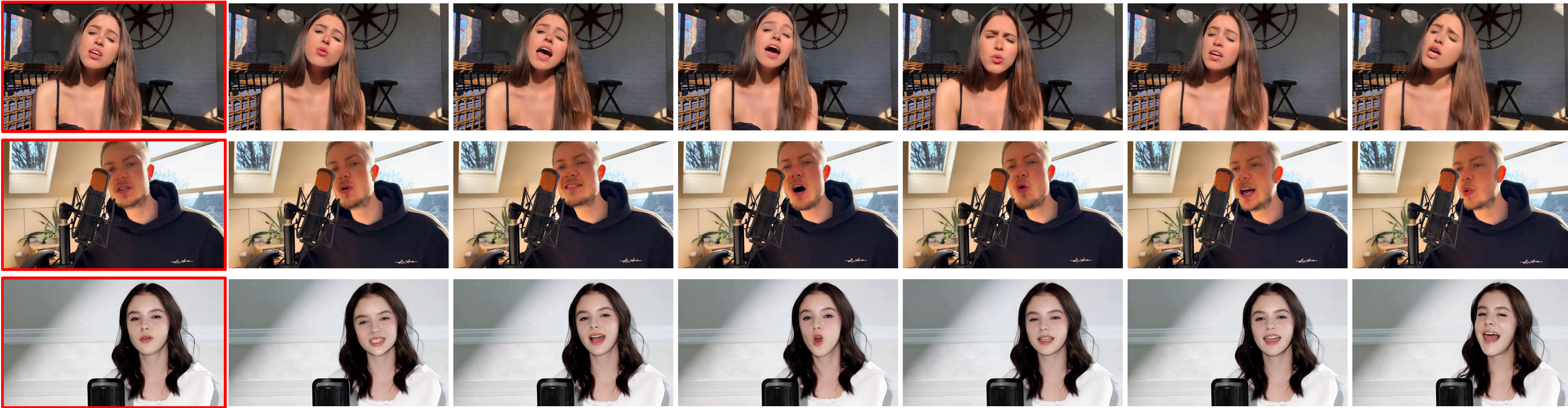}
\end{center}
\vspace{-0.55cm}
   \caption{Audio-driven long avatar video results (3/5). The images with red borders are the reference images.}
\label{fig:long_video_3}
\vspace{-0.5cm}
\end{figure*}

\begin{figure*}[t!]
\begin{center}
\includegraphics[width=1\linewidth]{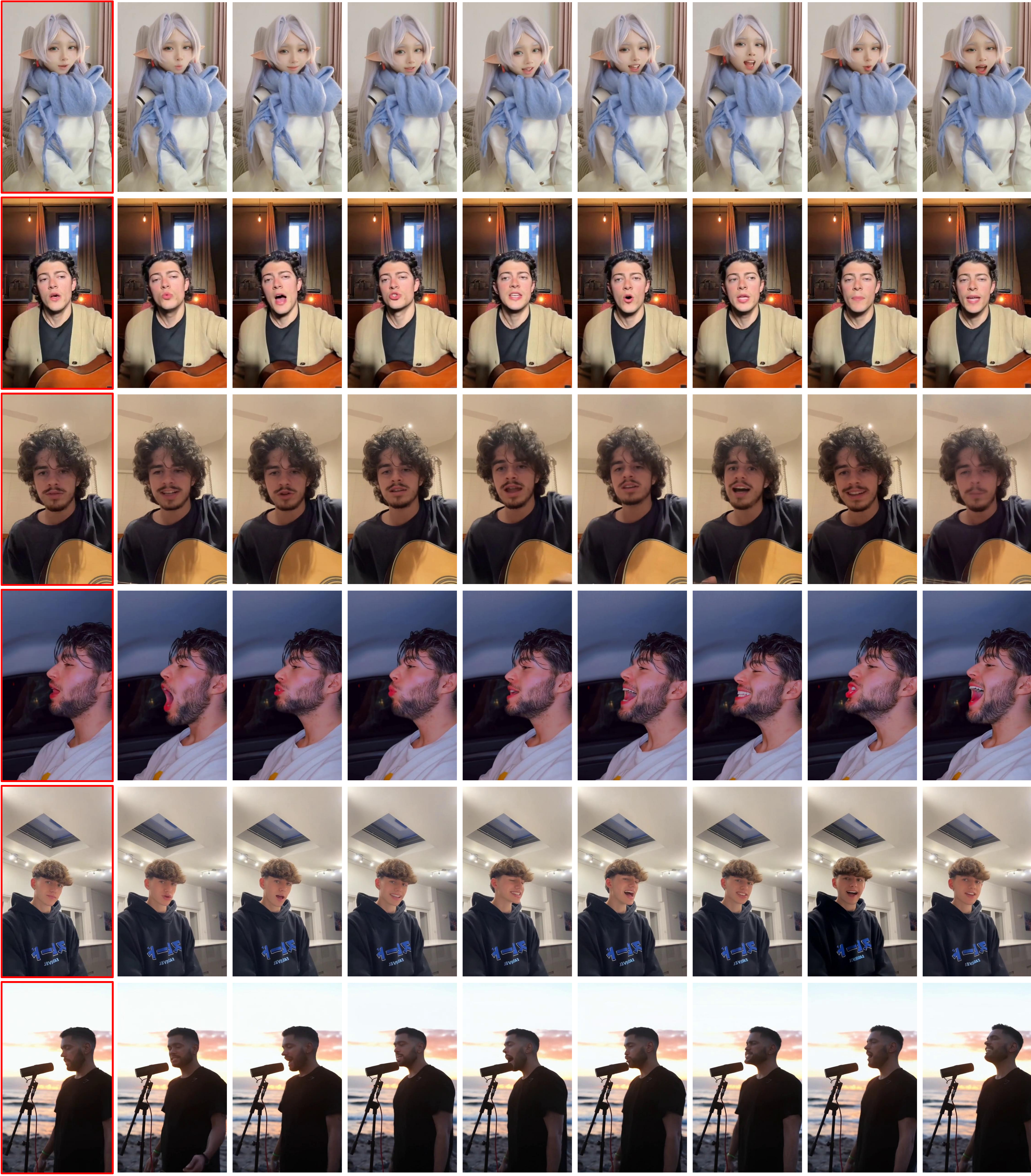}
\end{center}
\vspace{-0.55cm}
   \caption{Audio-driven long avatar video results (4/5). The images with red borders are the reference images.}
\label{fig:long_video_4}
\vspace{-0.5cm}
\end{figure*}

\begin{figure*}[t!]
\begin{center}
\includegraphics[width=1\linewidth]{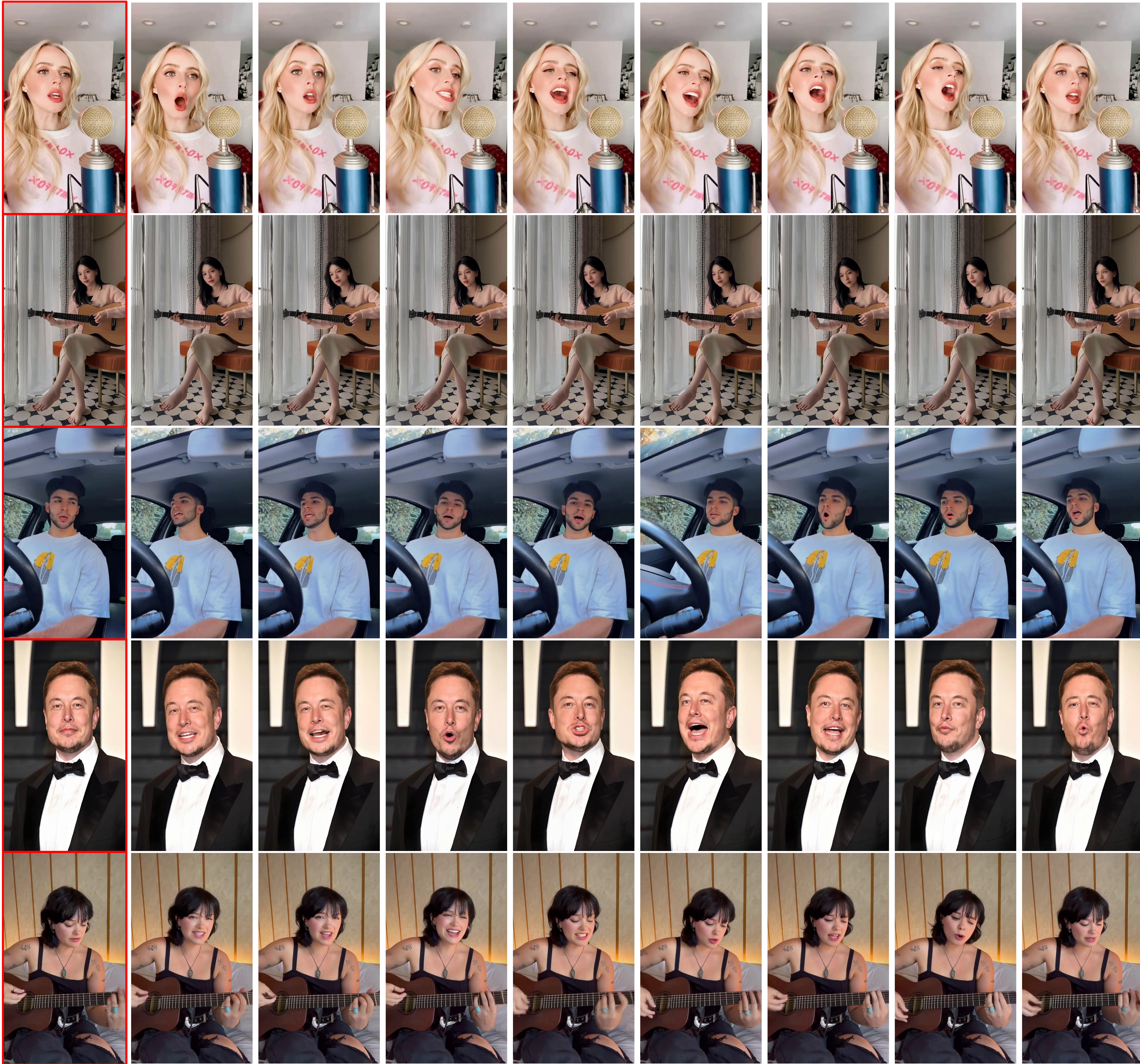}
\end{center}
\vspace{-0.55cm}
   \caption{Audio-driven long avatar video results (5/5). The images with red borders are the reference images.}
\label{fig:long_video_5}
\vspace{-0.5cm}
\end{figure*}

Fig. \ref{fig:long_video_1}, Fig. \ref{fig:long_video_2}, Fig. \ref{fig:long_video_3}, Fig. \ref{fig:long_video_4}, and Fig. \ref{fig:long_video_5} show additional audio-driven long avatar videos synthesized by our StableAvatar.
\textbf{Notably, each audio lasts over 3 minutes, and with an FPS of 30, this results in 5400+ synthesized frames. We only display selected frames from the last 10 seconds for brevity.}
Each case contains complex protagonist's appearance and an intricate audio rhythm pattern.
We can see that our StableAvatar can perform a wide range of audio-driven avatar animation while simultaneously preserving the protagonist’s appearance, background, and identity, even after synthesizing 5000+ frames.
For example, in the reference image in the third row of Fig. \ref{fig:long_video_1}, the protagonist wears a mask, making it particularly challenging for the model to perform accurate lip synchronization with audio.
Moreover, the reference protagonists in the fourth row and the sixth row of Fig. \ref{fig:long_video_4} face the camera at an angle, making it dramatically difficult for the avatar animation model to preserve ID consistency. 
It is noticeable that our StableAvatar can accurately manipulate the lip, facial expression, and body gesture in the reference image while preserving high-quality identity consistency, even in specific cases involving head shaking and external object interaction.
Moreover, StableAvatar supports simultaneously animating the avatar and background following the given audio via text prompt description.
For example, in the reference images in the second row of Fig. \ref{fig:long_video_1} and the sixth row of Fig. \ref{fig:long_video_4}, StableAvatar utilizes text prompts to animate contextual elements—for instance, making the scenery outside a car window shift as the vehicle starts moving, or generating dynamic ocean waves along the coastline.

\subsection{Limitation and Future Work}
\begin{figure*}[t!]
\begin{center}
\includegraphics[width=1\linewidth]{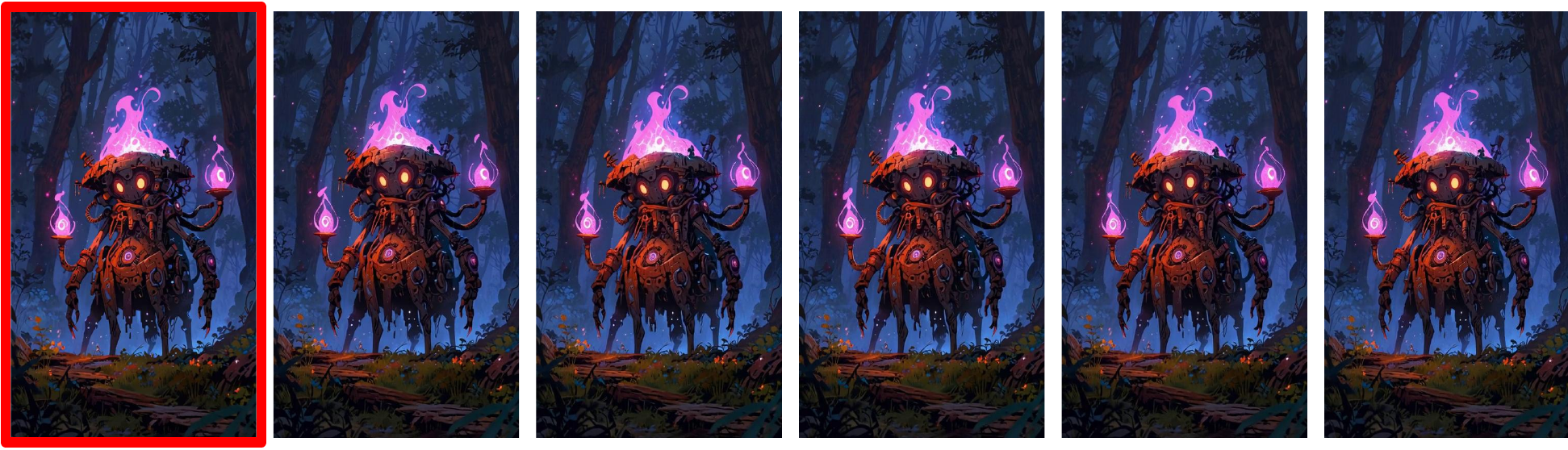}
\end{center}
\vspace{-0.55cm}
   \caption{One failure case of our StableAvatar. The images with red borders are the reference images.}
\label{fig:limitation}
\vspace{-0.5cm}
\end{figure*}

Fig. \ref{fig:limitation} shows one failure case of our StableAvatar. In the given reference image, the protagonist is a non-human creature (fantastical creature), which has significant appearance and structure discrepancies with the common human. Our StableAvatar struggles to find its lip, thereby failing to synchronize its lip with the audio.
One potential solution is to introduce an additional reference network to explicitly capture the semantic details of the reference images. This part is left as future work. 

\subsection{Ethical Concern}
Our StableAvatar can animate the given reference image based on the given audio, which can be implemented in various fields, including virtual reality and digital human creation. However, the potential misuse of this model, particularly for creating misleading content on social media platforms, is a concern. To mitigate this, it is essential to use sensitive content detection algorithms.

\end{document}